\documentclass{article} 
\usepackage{iclr2027_conference,times}

\usepackage{amsmath,amsfonts,bm}

\def\eqref#1{equation~\ref{#1}}

\def\1{\bm{1}}

\DeclareMathAlphabet{\mathsfit}{\encodingdefault}{\sfdefault}{m}{sl}
\SetMathAlphabet{\mathsfit}{bold}{\encodingdefault}{\sfdefault}{bx}{n}

\usepackage{hyperref}
\usepackage{url}
\usepackage{amsmath}
\usepackage{booktabs}
\usepackage{array}
\usepackage{graphicx}
\usepackage{wrapfig}
\usepackage{tabularx}
\usepackage{needspace}
\usepackage{colortbl}
\usepackage{flafter}

\definecolor{protttmain}{RGB}{226,238,250}
\definecolor{protttlight}{RGB}{241,247,252}

\title{ProTTT: Learning to Learn Semantic User Memory with Test-Time Training}

\author{
Sejun Park \quad Hyoungjo Bhang \quad Hyein Jung \quad
Yohan Jo\thanks{Corresponding author: \texttt{yohan.jo@snu.ac.kr}} \\
Graduate School of Data Science, Seoul National University \\
\texttt{\{aprimelonge,hyoungjo.bhang,hye\_\_n,yohan.jo\}@snu.ac.kr}
}

\iclrfinalcopy

\begin{document}
\raggedbottom
\maketitle
\lhead{Preprint}

\begin{abstract}
Personalization requires language models to capture user-specific knowledge from a growing user history. Existing context-based approaches incur increasing inference costs as user history accumulates and rely on separate retrieval or summarization stages, while parametric-based approaches often require reconstructing user representations when new user data is added. We introduce \textsc{ProTTT}, a profile-supervised meta-learning framework for learning semantic user memory. The memory construction starts from a shared initialization and is updated for each user through test-time training on user history, allowing it to evolve continuously as the history grows. However, since test-time training alone does not explicitly encourage the memory to capture semantic user knowledge necessary for personalization, we learn this shared initialization using textual user profiles as supervision, so that test-time training on user history captures semantic knowledge more effectively. \textsc{ProTTT} consistently outperforms both full history ICL and all parametric baselines across diverse benchmarks, while substantially reducing inference cost by compressing user history into a lightweight parameterized memory. Our analysis also shows that profile supervision is a reliable objective for learning semantic user knowledge and that the resulting memory can track and retain evolving user preferences, while remaining robust across different history sizes. Overall, we demonstrate the effectiveness of test-time training for personalization and establish \textsc{ProTTT} as a baseline for continuously evolving user memory. 
\end{abstract}

\section{Introduction}
\label{sec:introduction}

\begin{figure}[t]
\centering
\includegraphics[
width=\linewidth,
keepaspectratio
]{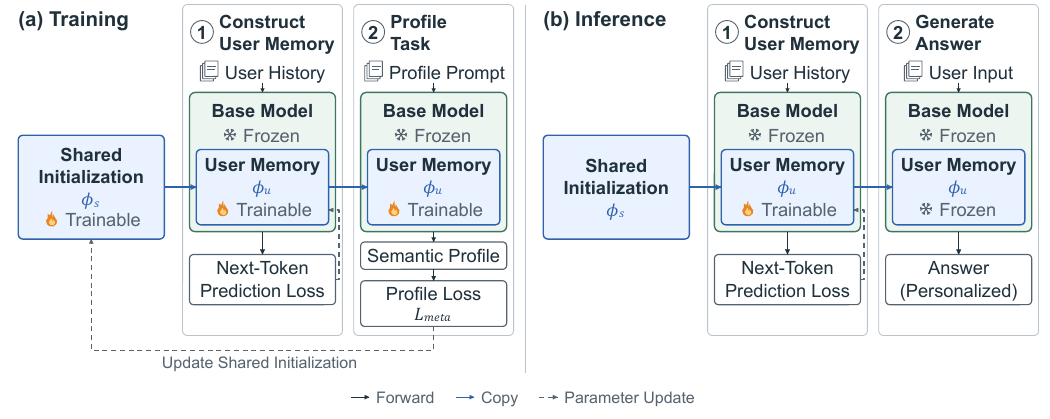}
\caption{
\textbf{Overview of \textsc{ProTTT}.} (a) During training, a shared initialization $\phi_s$ is adapted to each user through next-token prediction on the user history, producing user-specific memory $\phi_u$. The adapted memory is then evaluated on a semantic profile task, whose loss is backpropagated through the adaptation process to update $\phi_s$. (b) At inference time, the learned initialization is adapted using only the target user’s history and then frozen for personalized generation. 
}
\label{fig:framework-overview}
\end{figure}

As large language models (LLMs) are more deeply integrated into users' everyday workflows, it has become increasingly important to understand each user’s context and provide personalized responses. However, effective personalization often requires more than memorizing explicit information contained in the user’s data. The model must also capture implicit knowledge about the user---including preferences and behavioral tendencies---which is only revealed across a sequence of user content or utterances \citep{salemi-etal-2024-lamp, jiang-etal-2025-know}. We refer to this inferred, user-specific knowledge as \textit{semantic user memory}, and the sequence of user data as \textit{user history}. 

Existing personalization approaches can be classified into two categories. First, context-based approaches provide the user’s full history, retrieve a subset of relevant records, or summarize the history into a textual profile. While using the full history allows direct access to all user records, its inference cost scales with the history length and is bounded by the context window \citep{richardson-etal-2023-integrating}. Retrieval reduces the amount of user history included in the prompt, but personalization performance can vary substantially depending on which records are retrieved. Apart from its sensitivity to record selection, it also incurs an additional inference-time overhead \citep{salemi-etal-2024-lamp, mysore-etal-2024-pearl}. Summarized profiles can reduce the context length, but their effectiveness depends on the quality of the summary, and the profile must be revised as the user history grows \citep{richardson-etal-2023-integrating, zhang-2024-guided}.

Second, parameter-based approaches have been proposed that encode user history into learned representations, such as user-specific memory modules, prefix embeddings, or hypernetwork-generated parameters \citep{tan-etal-2024-democratizing, liu-etal-2025-llms, tan-etal-2026-instant}. These approaches show that compact representations can successfully encode user memory for personalization. However, as the user history grows, the representations must be updated to incorporate the new user data. Existing methods typically reprocess the entire history to construct an updated user representation. This motivates a new approach to user memory that can be updated continuously from its existing representation.

To that end, we introduce \textsc{\textbf{ProTTT}}, a profile-supervised test-time training framework for learning semantic user memory (Figure~\ref{fig:framework-overview}). We represent semantic user memory as a lightweight adapter attached to a frozen base language model and specialize it for each user through test-time training (TTT). During test-time training, the memory is updated through next-token prediction on the user history, naturally supporting continual and incremental updates as the user history accumulates. However, next-token prediction alone does not explicitly encourage the memory to extract high-level semantic knowledge for personalization. Therefore, \textsc{ProTTT} learns an initialization for the memory adapter using textual user profiles as supervision, so that the memory can capture semantic user knowledge more effectively when updated through test-time training on user history. We refer to this process as \textit{profile-supervised meta-learning}, since it optimizes the initialization to better learn user semantics. Note that profile supervision is used only during training to learn the initialization. At inference time, the memory is updated through test-time training on the user’s history alone, and the resulting user-specific memory is used to generate personalized responses to user queries. 

\textsc{ProTTT} is evaluated across diverse personalization settings, including personalized persuasiveness prediction on ChangeMyView (CMV) data, product-rating prediction on LaMP-3, and preference-aligned recommendation and generalization to new scenarios on PersonaMem Types~6 and~7 \citep{park-etal-2026-learning, salemi-etal-2024-lamp, jiang-etal-2025-know}. \textsc{ProTTT} consistently outperforms both full-history ICL and all parametric baselines across all benchmarks. Because \textsc{ProTTT} compresses the user history into a lightweight parameterized memory, it also significantly reduces inference cost, achieving a 10$\times$ reduction in latency compared with full-history ICL on CMV. 

We further examine \textsc{ProTTT}, yielding the following findings. First, among different meta-learning objectives, user profile prediction yields the most consistent improvements across all benchmarks, supporting its effectiveness for learning semantic user knowledge. Second, test-time training reflects recent history more strongly, allowing the learned memory to better track evolving user preferences. Interestingly, the memory also retains knowledge from earlier history more robustly than full-history ICL as the history grows, whereas ICL becomes progressively less effective at preserving older user knowledge. Third, \textsc{ProTTT} maintains relatively stable personalization performance across varying user-history sizes. Finally, reducing the memory scaling factor that controls the contribution of the memory to the residual stream helps preserve general capabilities while largely preserving personalization knowledge.

Our results demonstrate the potential of test-time training as a practical lightweight mechanism for constructing and incrementally updating semantic user memory. We hope this work provides a foundation for developing continuously evolving user memory in personalized language models.

\section{Related Work}
\label{sec:related-work}

\paragraph{Personalization.} Prior work on language model personalization can be broadly grouped into context-based and parameter-based approaches. First, context-based approaches supply the user data directly in the prompt at inference time. LaMP studies lexical, semantic, and time-aware retrieval across personalization tasks~\citep{salemi-etal-2024-lamp}, while subsequent works improve retrieval using feedback from the downstream model \citep{salemi-etal-2024-optimization, mysore-etal-2024-pearl}. Other approaches compress user history into textual profiles by constructing task-aware summaries, extracting distinctive user attributes, or jointly learning retrieval and summarization \citep{richardson-etal-2023-integrating, zhang-2024-guided, park-etal-2026-learning}. 

Second, parameter-based approaches instead encode user data into learned representations. OPPU learns a separate parameter-efficient module for each user, while Personalized Pieces constructs user-specific modules from parameter fragments shared across users \citep{tan-etal-2024-democratizing, tan-etal-2024-personalized}. P2P generates user-specific LoRA parameters from a constructed user profile through a hypernetwork \citep{hu2022lora, tan-etal-2026-instant}. Other methods represent users through learned embeddings while keeping the base language model frozen \citep{ning2024userllm, liu-etal-2025-llms}.

\paragraph{Test-Time Training.} Test-time training (TTT) adapts model parameters on test inputs using self-supervised objectives \citep{sun2020testtime}. Recent work extends this idea by treating inference-time updates as a form of memory: TTT layers use online-trained models as recurrent hidden states, while Titans introduces a neural long-term memory module that is updated during inference \citep{sun2025learning, behrouz2025titans}. More closely related to our setting, TTT-E2E meta-learns an initialization for next-token-prediction updates over long contexts, and PERK writes contextual information into low-rank adapter parameters for long-context reasoning \citep{tandon2025endtoend, chen2026perk}. GradMem similarly writes context into a compact prefix memory through gradient-based reconstruction, whereas $\delta$-mem maintains a fixed-size parametric state through online delta updates \citep{kuratov2026gradmem, lei2026deltamem}. Building on this view of test-time parameter updates as memory, \textsc{ProTTT} studies their use for personalization and introduces profile-supervised meta-learning to shape these updates toward semantic user information that transfers across downstream personalization tasks.

\section{Method}
\label{sec:method}

\begin{wrapfigure}[17]{r}{0.40\textwidth}
    \vspace{-3.7\baselineskip}
    \centering
    \setlength{\abovecaptionskip}{2pt}
    \setlength{\belowcaptionskip}{0pt}
    \includegraphics[
        width=\linewidth,
        keepaspectratio
    ]{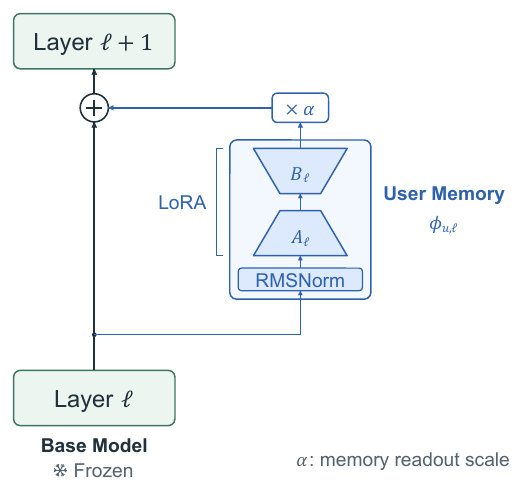}
    \caption{
    \textbf{User memory architecture.}
    A low-rank adapter reads the hidden state and adds its output to the residual stream.
    }
    \label{fig:parametric-memory}
    \vspace{-0.6\baselineskip}
\end{wrapfigure}

This section presents \textsc{ProTTT}, our profile-supervised test-time training framework for learning semantic user memory. We first define the adapter-based memory architecture, then describe how its initialization is meta-learned through profile supervision, and finally explain how the memory is built and used at inference time. 

\subsection{Parametric User Memory}
\label{sec:parametric-memory}

We implement user memory as lightweight additive adapter modules inserted between consecutive transformer layers. As illustrated in Figure~\ref{fig:parametric-memory}, each memory module reads the hidden representation produced by layer $\ell$, applies normalization followed by a low-rank transformation, and adds the resulting memory signal to the residual stream before the next layer. For a hidden state $z_\ell \in \mathbb{R}^{d}$,

\begin{equation}
    \tilde{z}_\ell = z_\ell + \alpha B_\ell A_\ell \operatorname{Norm}(z_\ell),
\end{equation}

where $A_\ell \in \mathbb{R}^{r \times d}$ and $B_\ell \in \mathbb{R}^{d \times r}$ are low-rank projection matrices with $r \ll d$. The resulting representation $\tilde{z}_\ell$ is then passed to layer $\ell+1$. The collection of adapter parameters across layers constitutes the user memory $\phi_u$.

The insertion design in between the layers allows the memory to be written into a separate set of parameters without modifying the layer parameters. The scalar $\alpha$ controls how strongly the memory output is added to the residual stream. We set $\alpha=1$ throughout training and use it as a memory readout scale at inference time.

\subsection{Training}
\label{sec:training}

As illustrated in Figure \ref{fig:framework-overview}(a), the training process consists of two parts: constructing user memory through test-time training and profile-supervised meta-learning for learning the shared initialization $\phi_s$. For each user, the memory construction begins with the shared initialization $\phi_s$. The memory parameters are then updated on the user history by next-token prediction. We first describe this memory construction process and then explain how profile supervision is used to learn $\phi_s$.

\paragraph{Constructing User Memory.} Given the history of the user $u$, $H_u = \{h_{u,1},\ldots,h_{u,n}\}$, where $h_{u,i}$ denotes the $i$-th historical record, we divide $H_u$ into chronologically ordered chunks $\{C_{u,1},C_{u,2},\ldots\}$. Starting from the shared initialization $\phi_s$, the memory is updated sequentially on each chunk using the next-token prediction objective. Specifically, we initialize $\phi_u^{(0)} \leftarrow \phi_s$ and update the memory after processing the $k$-th chunk as

\begin{equation}
\phi_u^{(k)}
\leftarrow
\phi_u^{(k-1)}
-
\eta_{\mathrm{ttt}}
\nabla
\mathcal{L}_{\mathrm{NTP}}
\left(C_{u,k}; \phi_u^{(k-1)}\right)
\label{eq:user-memory-construction}
\end{equation}

$\eta_{\mathrm{ttt}}$ denotes the learning rate for test-time training, and $\phi_u^{(k)}$ denotes the memory parameters after processing the $k$-th chunk. After all chunks have been processed, we denote the final user-specific memory as $\phi_u$. Additional implementation details are provided in Appendix~\ref{app:memory-writing-details}.

\paragraph{Profile-Supervised Meta-Learning.} Although test-time training naturally supports incremental updates as the user history grows, next-token prediction alone does not explicitly encourage memory to capture semantic user knowledge for personalization. We therefore introduce a meta-learning objective that uses textual user profiles as supervision to train the shared initialization $\phi_s$. The goal is to learn an initialization that helps test-time training construct semantic user memory more effectively.

For each training user $u$, let $x_u^{\mathrm{prof}}$ denote the profile-generation instruction and $y_u^{\mathrm{prof}}$ denote the corresponding semantic user profile. We first construct the user-specific memory $\phi_u$ from the user's history using the test-time training procedure described above. We then use the resulting memory to generate the user profile. The meta-learning loss is the standard next-token prediction loss over the target profile.

\begin{equation}
\mathcal{L}_{\mathrm{meta}}^{(u)}
=
-
\frac{1}{T_u}
\sum_{t=1}^{T_u}
\log
p_{\theta,\phi_u}
\left(
y_{u,t}^{\mathrm{prof}}
\mid
x_u^{\mathrm{prof}},
y_{u,<t}^{\mathrm{prof}}
\right).
\end{equation}

We normalize the loss by $T_u$, the number of tokens in the target profile, so that users with longer profiles do not contribute disproportionately to the meta-learning objective. The profile-generation loss is backpropagated through the sequence of test-time training updates used to construct $\phi_u$. This gradient is then used to update the shared initialization.

\begin{equation}
\phi_s
\leftarrow
\phi_s
-
\eta_{\mathrm{meta}}
\nabla_{\phi_s}
\mathcal{L}_{\mathrm{meta}}^{(u)}.
\end{equation}

$\eta_{\mathrm{meta}}$ denotes the meta-learning rate. Repeating this process across training users trains the shared initialization $\phi_s$ so that future test-time training can capture semantic user knowledge more effectively. %

\subsection{Inference}
\label{sec:inference}

Because the shared initialization $\phi_s$ is learned during training, inference requires only the target user's history, without user profile supervision or the meta-learning objective. For each target user $u$, we construct the user-specific memory $\phi_u$ by starting from $\phi_s$ and sequentially updating the memory on the user's history using the next-token prediction objective, following Equation~\ref{eq:user-memory-construction}. The resulting memory $\phi_u$ is then fixed and used to generate personalized responses.

As new user data accumulates, the memory can be updated directly from its current state without reprocessing the previous history. Given a new history chunk $C_{u,i}$, the memory is updated as

\begin{equation}
\phi_u'
\leftarrow
\phi_u
-
\eta_{\mathrm{ttt}}
\nabla
\mathcal{L}_{\mathrm{NTP}}
\left(
C_{u,i}; \phi_u
\right)
\end{equation}

This incremental update allows the memory to incorporate new user data as the history grows without repeatedly processing previously observed history.

\section{Experiments}
\label{sec:experiments}

We evaluate \textsc{ProTTT} across multiple personalization benchmarks covering persuasion, rating prediction, recommendation, and generalization. We first describe the experimental setup and the baselines, then report personalization performance and inference efficiency.

\subsection{Experimental Setup}
\label{sec:experimental-setup}

\paragraph{Training.} We use Qwen3-0.6B as the base language model in all experiments. For profile supervision, we use Tasks 1, 2, and 4 of ALPSBench \citep{xiao-etal-2026-alpsbench}, together with semantic profiles constructed from the histories of CMV training users. The implementation details are provided in Appendix~\ref{app:training-details}.

\paragraph{Benchmarks.} We evaluate \textsc{ProTTT} on CMV, LaMP-3, and PersonaMem. CMV and LaMP-3 evaluate personalized persuasiveness and rating prediction. For PersonaMem, we focus on Types~6 and~7, which assess preference-aligned recommendation and generalization of user knowledge to new scenarios. These tasks most directly evaluate whether a model has formed semantic user knowledge that can guide behavior beyond explicit recall of the user history. The results of other PersonaMem types are reported in Appendix~\ref{app:additional-personamem}.

\paragraph{Baselines.} The in-context baselines include full history, recent-5, BM25 retrieval, profile-augmented generation (PAG), and PAG with the full history \citep{richardson-etal-2023-integrating}. For parameter-based baselines, we include user-specific LoRA following the setting in PRIME \citep{zhang-etal-2025-prime}, P2P, which uses a shared hypernetwork \citep{tan-etal-2026-instant}, and $\delta$-Mem, which encodes context into a parameter state \citep{lei2026deltamem}. The implementation details are provided in Appendix~\ref{app:baseline-details}.

\paragraph{Efficiency.} We measure efficiency using a single NVIDIA H200 NVL. The main table reports representative measurements on CMV; complete results for CMV, LaMP-3, and PersonaMem, together with the measurement details, are provided in Appendix~\ref{app:efficiency-details}.

\subsection{Personalization Performance and Efficiency}
\label{sec:personalization-performance-efficiency}

\begin{table}[t]
\centering
\caption{
\textbf{Personalization performance and efficiency.}
\textit{w/o user} denotes the corresponding method without target-user
information. For \textsc{ProTTT}, this denotes the shared initialization
before constructing memory from the target user's history; $\delta$-Mem uses
an empty memory state, whereas P2P uses an empty user profile and history.
Bold indicates the best personalization result in each column.
$^\dagger$PAG profiles are generated offline using GPT-5-mini through an
external API; profile-construction latency is therefore excluded from the
H200 measurements.
\vspace{0.1cm}
}
\label{tab:main-results}
\setlength{\tabcolsep}{3.0pt}
\renewcommand{\arraystretch}{1.06}
\resizebox{\linewidth}{!}{%
\begin{tabular}{
@{}l
rrrrr
!{\vrule width 0.8pt}
rrr
@{}
}
\toprule
&
\multicolumn{5}{c}{\textbf{Personalization}}
&
\multicolumn{3}{c}{\textbf{Efficiency on CMV}}
\\
\cmidrule(lr){2-6}
\cmidrule(lr){7-9}
\textbf{Method}
& \shortstack{\textbf{CMV}\\AUC $\uparrow$}
& \shortstack{\textbf{LaMP-3}\\MAE $\downarrow$}
& \shortstack{\textbf{LaMP-3}\\RMSE $\downarrow$}
& \shortstack{\textbf{PMem}\\T6 $\uparrow$}
& \shortstack{\textbf{PMem}\\T7 $\uparrow$}
& \shortstack{\textbf{History Proc.}\\(s/user) $\downarrow$}
& \shortstack{\textbf{Prompt}\\(tokens/query) $\downarrow$}
& \shortstack{\textbf{Latency}\\(ms/query) $\downarrow$}
\\
\midrule
Base
& .466 & 2.103 & 2.602 & .273 & .088
& -- & 748 & 15.3
\\
\midrule
\multicolumn{9}{l}{\textit{In-Context Baselines}}
\\
\addlinespace[1pt]

Full History
& .475 & .892 & 1.569 & .273 & .140
& -- & 18.9K & 273.5
\\
RAG (Recent-5)
& .456 & 1.800 & 2.434 & .418 & .140
& -- & 2.2K & 20.0
\\
RAG (BM25)
& .436 & 1.763 & 2.405 & \textbf{.491} & .246
& -- & 4.3K & 39.2
\\
PAG
& .411 & 2.981 & 3.282 & .309 & .123
& n/a$^\dagger$ & 1.1K & 15.9
\\
PAG + Full History
& .494 & 1.299 & 2.025 & .273 & .123
& n/a$^\dagger$ & 19.2K & 278.1
\\
\midrule
\multicolumn{9}{l}{\textit{Parametric Baselines}}
\\
\addlinespace[1pt]

User-Specific LoRA
& .524 & 1.804 & 2.460 & .236 & .088
& 11.74 & 748 & 29.6
\\
\addlinespace[2pt]

P2P
& .570 & 1.216 & 1.571 & .255 & .088
& 5.43 & 748 & 84.7
\\
\quad w/o user
& .571 & 1.236 & 1.594 & .255 & .088
& -- & -- & --
\\
\addlinespace[2pt]

\(\delta\)-Mem
& .554 & 1.064 & 1.354 & .309 & .140
& .29 & 748 & 31.0
\\
\quad w/o user
& .476 & 2.187 & 2.673 & .309 & .088
& -- & -- & --
\\
\midrule
\multicolumn{9}{l}{\textit{Training Control}}
\\
\addlinespace[1pt]

TTT (naive)
& .501 & 2.475 & 2.864 & .273 & .140
& -- & -- & --
\\
\midrule
\multicolumn{9}{l}{\textit{Ours}}
\\
\addlinespace[1pt]

\rowcolor{protttmain}
\textsc{ProTTT}
& \textbf{.624} & \textbf{.767} & \textbf{1.112}
& .327 & \textbf{.298}
& 2.82 & 748 & 26.2
\\
\rowcolor{protttlight}
\quad w/o user
& .518 & 1.346 & 1.548
& .218 & .246
& -- & -- & --
\\
\bottomrule
\end{tabular}%
}
\end{table}

\paragraph{In-Context Baselines.}
Table~\ref{tab:main-results} reports personalization performance together with inference efficiency. Among the in-context baselines, PAG + Full History performs best on CMV, while using the full history is most effective on LaMP-3. Retrieval-based methods perform better on PersonaMem, with RAG (BM25) achieving the best result on Type~6. In contrast, we observe that adding a semantic user profile alone degrades performance on several tasks relative to the base model. When combined with the full user history, the profile improves CMV performance but degrades LaMP-3 and PersonaMem Type~7. It is notable that providing semantic user information explicitly as text does not necessarily improve personalization, suggesting that in-context learning may not make effective use of such information even when it is directly available in the prompt.

\paragraph{Parametric Baselines.} The parametric baselines generally improve over the base model, although the magnitude varies across methods. Independently fitting a user-specific LoRA yields relatively smaller gains, whereas $\delta$-Mem achieves greater and consistent improvements over the base model. It even outperforms the full history baseline on CMV and PersonaMem, while remaining competitive with full history on LaMP-3 depending on the metric. While P2P shows the strongest performance on CMV, it is interesting to find that P2P produces similar scores with and without user-specific information, suggesting that much of its performance comes from task-level knowledge rather than user-specific details.

\paragraph{\textsc{ProTTT}.} Next, we evaluate TTT (naive), an ablation that applies the same test-time training procedure as \textsc{ProTTT} without profile supervision. It adapts to each user’s history following Equation~\ref{eq:user-memory-construction}, but produces inconsistent gains across tasks. On the other hand, our method \textsc{ProTTT}, which adds profile-supervised meta-learning, achieves the best performance on CMV, LaMP-3, and PersonaMem Type~7. Importantly, it also outperforms full history and all parametric baselines across all datasets. The results reveal that test-time training alone is insufficient, and that profile supervision is critical for guiding the adaptation toward effective semantic user memory.

\paragraph{Efficiency.}
We also examine the efficiency of the compared methods. Parametric methods substantially reduce prompt length by encoding user information into reusable parameters rather than repeatedly providing the full history. \textsc{ProTTT} retains this advantage while achieving stronger personalization performance. Concretely, it reduces the average prompt length $25.3\times$ from 18.9K to 748 tokens and per-query latency $10.4\times$ from 273.5 ms to 26.2 ms compared to full history on CMV. Constructing the user memory requires only a one-time history-processing cost of 2.82 seconds per user, after which it can be reused and amortized across subsequent queries.

\paragraph{\textsc{ProTTT} Across Backbone Models.}
We additionally apply \textsc{ProTTT} to Llama-3.2-1B-Instruct and separately meta-learn its memory initialization for this backbone. Across all reported evaluations, \textsc{ProTTT} outperforms full history ICL. These results show that the proposed memory architecture and profile-supervised meta-learning procedure remain effective when applied to a different language-model backbone. We provide complete results and analyses in Appendix~\ref{app:additional-backbone}.

\section{Analysis}
\label{sec:analysis}

\subsection{Which Meta-Learning Objective Best Guides User Memory Formation?}
\label{sec:meta-learning-objective-analysis}

\begin{figure}[h]
    \centering
    \includegraphics[width=\linewidth]{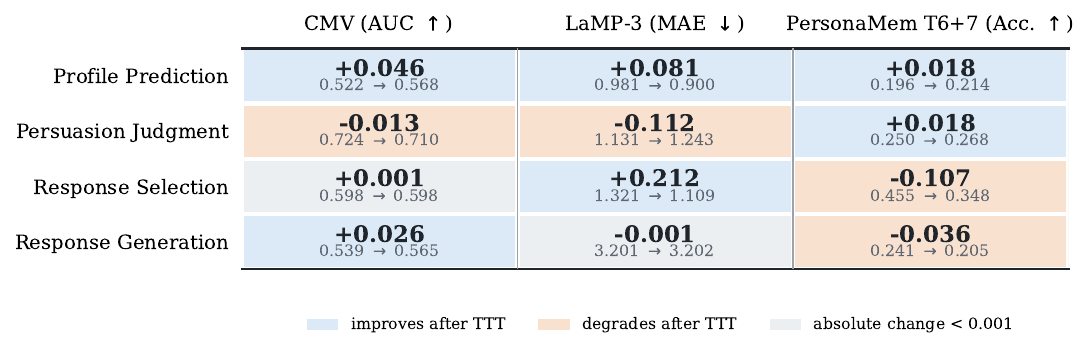}
    \caption{
    \textbf{Effect of the meta-learning objective on TTT.} The figure reports downstream task performance changes before and after user memory construction for each meta-learning objective. Bold values show the signed change, and smaller values show the corresponding raw scores. Positive values indicate improvement.
    }
    \label{fig:meta-learning-objective}
\end{figure}

We first analyze how the choice of meta-learning objective affects semantic user memory construction. We keep the test-time training procedure fixed and replace profile prediction with several alternative objectives: persuasion judgment, response selection, and response generation. The experimental details are provided in Appendix~\ref{app:meta-learning-objective-details}.

Figure~\ref{fig:meta-learning-objective} shows that profile prediction consistently improves performance across all three evaluation settings, whereas each alternative objective improves some tasks while degrading others. The results suggest that the meta-learning objective plays a critical role in determining what information test-time training captures. Profile supervision provides the most consistent signal across tasks, supporting our use of semantic user profiles as the meta-learning objective in \textsc{ProTTT}.

\subsection{Continual Acquisition and Retention of User Knowledge}
\label{sec:continual-memory-analysis}

\begin{figure}[h]
    \centering
    \includegraphics[width=\linewidth]{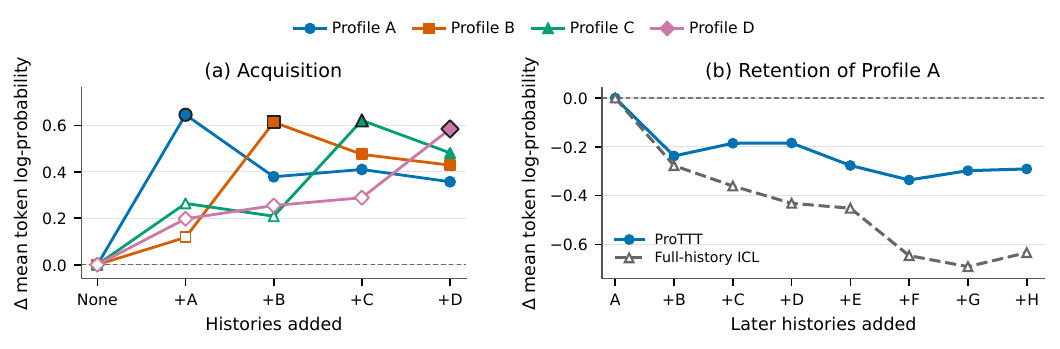}
    \caption{
    \textbf{Continual acquisition and retention of semantic user memory.} (a) Mean token log-probability of each gold user profile as profile-associated histories are sequentially incorporated. (b) Change in the log-probability of Profile~A as additional user histories are added, comparing \textsc{ProTTT} with full history ICL. Smaller degradation indicates better retention of previously acquired user information.
    }
    \label{fig:continual-user-memory}
\end{figure}

We next examine how semantic user memory evolves as new user data are incorporated over time. To simulate a growing user history, we expand it incrementally by adding one user history at a time, with experimental details provided in Appendix~\ref{app:continual-construction}. Figure~\ref{fig:continual-user-memory}(a) tracks the mean token log-probability of four gold user profiles as their corresponding histories are sequentially added through test-time training. For each profile, its mean token log-probability increases substantially once the associated history is incorporated, indicating that \textsc{ProTTT} can continually acquire new semantic user data.

The retention behavior is further explored in Figure~\ref{fig:continual-user-memory}(b), by tracking Profile~A as seven later user histories are incorporated sequentially. Both \textsc{ProTTT} and full history ICL gradually lose information about the earlier profile, but the degradation is considerably smaller for \textsc{ProTTT}. As the history grows, full history ICL progressively reduces the likelihood of Profile~A, whereas \textsc{ProTTT} maintains a relatively stable likelihood. These results suggest that test-time training not only incorporates newly arriving user information, but also retains previously acquired semantic knowledge more robustly as the user history accumulates.

\subsection{Scaling with User-History Size}
\label{sec:history-size-analysis}

\begin{figure}[!ht]
    \centering
    \includegraphics[width=\linewidth]{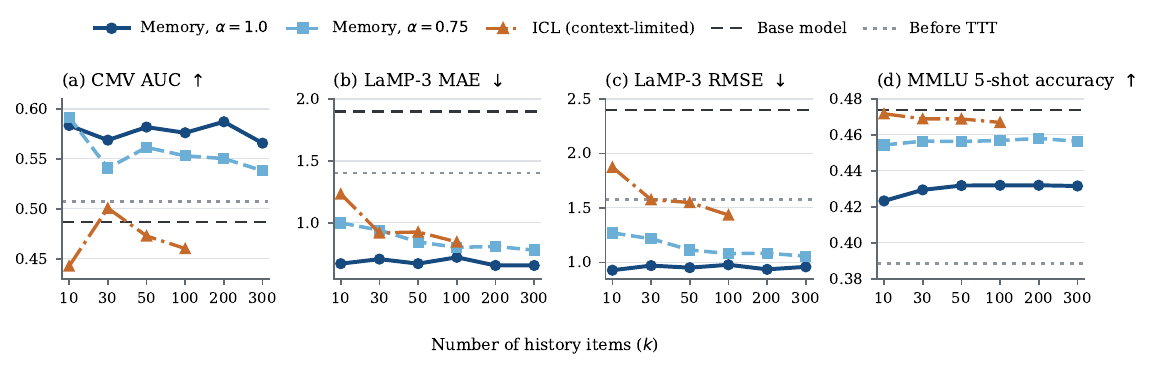}
    \caption{
    \textbf{Personalization and general capability as user history accumulates.}
    The plots show the scores after constructing user memory from CMV user histories.
    Curves compare \textsc{ProTTT} under different memory readout scales against full history ICL as the number of history items \(k\) increases.
    Horizontal lines denote the base model and the shared initialization before TTT.
    Full history ICL is reported only up to \(k=100\) because of the context limit.
    }
    \label{fig:history-size}
\end{figure}

We vary the number of history items used for personalization from $10$ to $300$, only using users with at least $300$ available items. TTT processes the most recent $k$ items, whereas ICL is evaluated only up to $k=100$ because of the context limit. We additionally evaluate 5-shot MMLU after constructing user memory at each history size for nine CMV users, allowing us to examine whether incorporating more user history progressively degrades the model's general capabilities.

As shown in Figure~\ref{fig:history-size}, TTT consistently outperforms ICL on CMV and LaMP-3 across the range where they can be compared. On CMV, ICL performance declines as $k$ increases from $30$ to $100$, whereas memory performance remains comparatively stable. Unlike ICL, TTT can process histories of up to $300$ items without increasing the context length at inference time, and its personalization performance shows no degradation as $k$ increases.

MMLU performance also shows no degradation as more history is incorporated. Full memory readout provides stronger personalization, while reducing $\alpha$ narrows the MMLU gap from the base model. Readout scaling therefore offers a direct trade-off between personalization and preserving the backbone model's general capability.

\subsection{General Capability, Generation Quality, and User-Information Recall}
\label{sec:generation-analysis}

\begin{table}[t]
\centering
\caption{
\textbf{General capability, generation quality, and user-information recall.}
(a) Reports 5-shot MMLU accuracy after processing $k=100$ history items and generation metrics on $40$ WildChat prompts.
(b) Reports the mean number of claims per generated user profile.
Sup., Gen., Unsup., and Contr. denote supported, generic, unsupported, and contradicted claims, respectively.
}
\label{tab:generation-quality}

\small
\setlength{\tabcolsep}{3pt}
\renewcommand{\arraystretch}{1.10}

\resizebox{\linewidth}{!}{%
\begin{tabular}{@{}lrrrrr|rrrrr@{}}
\toprule
& \multicolumn{5}{c|}{\textbf{(a) General capability and generation quality}}
& \multicolumn{5}{c}{\textbf{(b) Claims in generated user profiles}}
\\
\cmidrule(lr){2-6}
\cmidrule(l){7-11}
\textbf{Method}
& \shortstack{\textbf{MMLU}\\\textbf{Acc.} $\uparrow$}
& \shortstack{\textbf{Non-}\\\textbf{term.} $\downarrow$}
& \shortstack{\textbf{Seq.-}\\\textbf{rep.-2} $\downarrow$}
& \shortstack{\textbf{Distinct-2}\\$\uparrow$}
& \shortstack{\textbf{Self-}\\\textbf{BLEU} $\downarrow$}
& \shortstack{\textbf{Claims}\\\textbf{/gen.}}
& \textbf{Sup.}
& \textbf{Gen.}
& \textbf{Unsup.}
& \textbf{Contr.}
\\
\midrule

Base
& \textbf{.474} & .000 & .057 & .605 & .359
& 6.22 & 1.67 & 2.27 & 2.24 & .04
\\
\midrule

User-specific LoRA
& .397 & .740 & .417 & .246 & .463
& 5.11 & 1.24 & 1.49 & 2.13 & .24
\\

P2P
& .390 & .006 & .064 & .671 & .255
& 3.29 & .91 & .13 & 1.89 & .36
\\

$\delta$-Mem
& \textbf{.474} & .003 & .123 & .563 & .309
& 9.87 & 2.80 & 2.09 & 4.53 & .44
\\
\midrule

\textsc{ProTTT}, $\alpha=.75$
& .457 & .009 & .092 & .605 & .277
& 8.04 & \textbf{5.69} & 1.02 & 1.22 & .11
\\

\textsc{ProTTT}, $\alpha=1$
& .432 & .073 & .165 & .469 & .307
& 7.42 & 4.27 & .58 & 2.36 & .22
\\
\bottomrule
\end{tabular}%
}
\end{table}

We evaluate general capability using 5-shot MMLU, generation quality using held-out WildChat prompts \citep{zhao2024wildchat}, and user-information recall by asking the model to describe the user without providing the history in context. We compare \textsc{ProTTT} with user-specific LoRA and $\delta$-Mem as parametric reference methods. The implementation and baseline details are provided in Appendix~\ref{app:generation-analysis-details} and Appendix~\ref{app:baseline-details}.

Table~\ref{tab:generation-quality}(a) shows that the effect of parametric personalization on general model behavior varies substantially across methods. On MMLU, \textsc{ProTTT} with $\alpha=0.75$ achieves accuracy comparable to that of the base model, which is considerably smaller than the reduction observed with of user-specific LoRA. $\delta$-Mem remains at the base-model level on MMLU, but provides substantially weaker user-information recall, as shown in Table~\ref{tab:generation-quality}(b). Using \textsc{ProTTT} at full strength produces a larger MMLU reduction, indicating that the readout scale controls how strongly the user memory affects the model's general capabilities.

The WildChat results show a similar effect of readout scaling in generation. User-specific LoRA degrades severely, with $74\%$ of its responses reaching the generation limit and showing substantial increases in repetition. $\delta$-Mem largely avoids non-termination, but produces more repetition and lower lexical diversity than \textsc{ProTTT} with $\alpha=0.75$. Full-strength \textsc{ProTTT} also diverges from the base model's generation behavior, whereas the scaled readout keeps non-termination below $1\%$, maintains relatively low repetition, and preserves lexical diversity. Thus, scaling the memory readout substantially reduces both general-capability and generation-quality degradation.

Table~\ref{tab:generation-quality}(b) shows that this preservation does not come at the expense of access to the user information learned. At $\alpha=0.75$, \textsc{ProTTT} generates $5.69$ history-supported claims per response, compared to $1.67$ for the base model, $1.24$ for user-specific LoRA, and $2.80$ for $\delta$-Mem. It also produces fewer generic, unsupported, and contradicted claims than either parametric baseline. Although the full-strength readout still produces more supported user information than the baselines, it yields more unsupported claims than the scaled model. These results show that the scaled \textsc{ProTTT} readout provides the strongest overall balance between preserving general model behavior, maintaining generation quality, and expressing information supported by the user's history. Appendix~\ref{app:readout-scale-sweep} reports the full readout-scale sweep, $\alpha\in\{0.25,0.5,0.75,1.0\}$, for both Qwen3-0.6B and Llama-3.2-1B-Instruct, covering personalization, MMLU and WildChat generation quality.

\section{Conclusion}
\label{sec:conclusion}


We introduced \textsc{ProTTT}, a profile-supervised test-time training framework for constructing semantic user memory with lightweight adapters. \textsc{ProTTT} learns a shared initialization from textual user profiles and then constructs user-specific memory through test-time training on user history, enabling the memory to be updated continuously as the history grows. Across multiple personalization benchmarks, \textsc{ProTTT} consistently outperforms full history ICL and parametric baselines while reducing inference cost through a compact parameterized memory. Our analyses further show that profile supervision provides a particularly effective objective for learning semantic user knowledge and that the resulting memory can track and retrain evolving user preferences while remaining robust across different history sizes. We hope this work provides a useful foundation for future research on efficient, continuously evolving user memory for personalized language models.

\subsection*{AI use statement}
We used generative AI tools to assist with code development and debugging, literature search and review, and the drafting and language editing of this paper. GPT-5-mini was also used in our experiments to generate semantic-profile supervision from user histories, screen user profiles for compatibility in the continual-memory analysis, and classify claims in generated user descriptions. The authors reviewed and verified all AI-assisted code, text, citations, and experimental artifacts. The authors take responsibility for the final content of this work.



\subsection*{Ethics statement}
This work uses existing research datasets and does not collect new user data or conduct new human-subject experiments. Nevertheless, user histories can contain personal opinions, preferences, and other sensitive information. A system that internalizes such information may create risks of unauthorized profiling, privacy violations, inaccurate user representations, and manipulative personalization, particularly in persuasive applications. Practical deployment should therefore require user consent, appropriate protection of stored memory, and mechanisms that allow users to inspect, correct, and delete their information. The LLM-generated profiles and automatic judgments used in our experiments may also contain errors or biases and should not be treated as authoritative descriptions of users.


\subsection*{Reproducibility statement}
Section~\ref{sec:method} describes the memory architecture, TTT memory-writing procedure, profile-supervised meta-learning objective, and inference procedure. Section~\ref{sec:experimental-setup} describes the training data, evaluation benchmarks, and baselines. Appendices~\ref{app:training-details} and~\ref{app:memory-writing-details} provide data construction, training, history serialization, and memory-writing details. Baseline implementations and efficiency measurements are documented in Appendices~\ref{app:baseline-details} and~\ref{app:efficiency-details}. The construction of the meta-learning objectives, continual-memory streams, and generation evaluation is described in Appendices~\ref{app:meta-learning-objective-details}, \ref{app:continual-construction}, and~\ref{app:generation-analysis-details}. Additional PersonaMem results are reported in Appendix~\ref{app:additional-personamem}.



We will publicly release the source code upon acceptance.

\bibliography{iclr2027_conference}

@inproceedings{salemi-etal-2024-lamp,
  title     = {{L}a{MP}: When Large Language Models Meet Personalization},
  author    = {Salemi, Alireza and Mysore, Sheshera and Bendersky, Michael and Zamani, Hamed},
  booktitle = {Proceedings of the 62nd Annual Meeting of the Association for Computational Linguistics (Volume 1: Long Papers)},
  month     = aug,
  year      = {2024},
  address   = {Bangkok, Thailand},
  publisher = {Association for Computational Linguistics},
  pages     = {7370--7392},
  doi       = {10.18653/v1/2024.acl-long.399},
  url       = {https://aclanthology.org/2024.acl-long.399/}
}

@misc{richardson-etal-2023-integrating,
  title         = {Integrating Summarization and Retrieval for Enhanced Personalization via Large Language Models},
  author        = {Richardson, Chris and Zhang, Yao and Gillespie, Kellen and Kar, Sudipta and Singh, Arshdeep and Raeesy, Zeynab and Khan, Omar Zia and Sethy, Abhinav},
  year          = {2023},
  eprint        = {2310.20081},
  archivePrefix = {arXiv},
  primaryClass  = {cs.CL},
  url           = {https://arxiv.org/abs/2310.20081}
}

@inproceedings{mysore-etal-2024-pearl,
  title     = {{Pearl}: Personalizing Large Language Model Writing Assistants with Generation-Calibrated Retrievers},
  author    = {Mysore, Sheshera and Lu, Zhuoran and Wan, Mengting and Yang, Longqi and Sarrafzadeh, Bahareh and Menezes, Steve and Baghaee, Tina and Gonzalez, Emmanuel Barajas and Neville, Jennifer and Safavi, Tara},
  booktitle = {Proceedings of the 1st Workshop on Customizable NLP: Progress and Challenges in Customizing NLP for a Domain, Application, Group, or Individual (CustomNLP4U)},
  month     = nov,
  year      = {2024},
  address   = {Miami, Florida, USA},
  publisher = {Association for Computational Linguistics},
  pages     = {198--219},
  doi       = {10.18653/v1/2024.customnlp4u-1.16},
  url       = {https://aclanthology.org/2024.customnlp4u-1.16/}
}

@inproceedings{salemi-etal-2024-optimization,
  title     = {Optimization Methods for Personalizing Large Language Models through Retrieval Augmentation},
  author    = {Salemi, Alireza and Kallumadi, Surya and Zamani, Hamed},
  booktitle = {Proceedings of the 47th International ACM SIGIR Conference on Research and Development in Information Retrieval},
  series    = {SIGIR '24},
  year      = {2024},
  pages     = {752--762},
  publisher = {Association for Computing Machinery},
  doi       = {10.1145/3626772.3657783},
  url       = {https://doi.org/10.1145/3626772.3657783}
}

@inproceedings{zhang-2024-guided,
  title     = {Guided Profile Generation Improves Personalization with Large Language Models},
  author    = {Zhang, Jiarui},
  booktitle = {Findings of the Association for Computational Linguistics: EMNLP 2024},
  month     = nov,
  year      = {2024},
  address   = {Miami, Florida, USA},
  publisher = {Association for Computational Linguistics},
  pages     = {4005--4016},
  doi       = {10.18653/v1/2024.findings-emnlp.231},
  url       = {https://aclanthology.org/2024.findings-emnlp.231/}
}

@inproceedings{park-etal-2026-learning,
  title     = {Learning to Retrieve User History and Generate User Profiles for Personalized Persuasiveness Prediction},
  author    = {Park, Sejun and Park, Yoonah and Lim, Jongwon and Jo, Yohan},
  booktitle = {Findings of the Association for Computational Linguistics: ACL 2026},
  month     = jul,
  year      = {2026},
  address   = {San Diego, California, United States},
  publisher = {Association for Computational Linguistics},
  pages     = {17338--17359},
  doi       = {10.18653/v1/2026.findings-acl.858},
  url       = {https://aclanthology.org/2026.findings-acl.858/}
}

@inproceedings{jiang-etal-2025-know,
  title     = {Know Me, Respond to Me: Benchmarking {LLM}s for Dynamic User Profiling and Personalized Responses at Scale},
  author    = {Jiang, Bowen and Hao, Zhuoqun and Cho, Young-Min and Li, Bryan and Yuan, Yuan and Chen, Sihao and Ungar, Lyle and Taylor, Camillo J. and Roth, Dan},
  booktitle = {Second Conference on Language Modeling},
  year      = {2025},
  url       = {https://openreview.net/forum?id=6ox8XZGOqP}
}

@inproceedings{tan-etal-2024-democratizing,
  title     = {Democratizing Large Language Models via Personalized Parameter-Efficient Fine-tuning},
  author    = {Tan, Zhaoxuan and Zeng, Qingkai and Tian, Yijun and Liu, Zheyuan and Yin, Bing and Jiang, Meng},
  booktitle = {Proceedings of the 2024 Conference on Empirical Methods in Natural Language Processing},
  month     = nov,
  year      = {2024},
  address   = {Miami, Florida, USA},
  publisher = {Association for Computational Linguistics},
  pages     = {6476--6491},
  doi       = {10.18653/v1/2024.emnlp-main.372},
  url       = {https://aclanthology.org/2024.emnlp-main.372/}
}

@inproceedings{hu2022lora,
  title     = {Lo{RA}: Low-Rank Adaptation of Large Language Models},
  author    = {Hu, Edward J. and Shen, Yelong and Wallis, Phillip and Allen-Zhu, Zeyuan and Li, Yuanzhi and Wang, Shean and Wang, Lu and Chen, Weizhu},
  booktitle = {International Conference on Learning Representations},
  year      = {2022},
  url       = {https://openreview.net/forum?id=nZeVKeeFYf9}
}

@inproceedings{tan-etal-2024-personalized,
  title     = {Personalized Pieces: Efficient Personalized Large Language Models through Collaborative Efforts},
  author    = {Tan, Zhaoxuan and Liu, Zheyuan and Jiang, Meng},
  booktitle = {Proceedings of the 2024 Conference on Empirical Methods in Natural Language Processing},
  month     = nov,
  year      = {2024},
  address   = {Miami, Florida, USA},
  publisher = {Association for Computational Linguistics},
  pages     = {6459--6475},
  doi       = {10.18653/v1/2024.emnlp-main.371},
  url       = {https://aclanthology.org/2024.emnlp-main.371/}
}

@inproceedings{tan-etal-2026-instant,
  title     = {Instant Personalized Large Language Model Adaptation via Hypernetwork},
  author    = {Tan, Zhaoxuan and Zhang, Zixuan and Wen, Haoyang and Li, Zheng and Zhang, Rongzhi and Chen, Pei and Mo, Fengran and Liu, Zheyuan and Zeng, Qingkai and Yin, Qingyu and Jiang, Meng},
  booktitle = {Proceedings of the 64th Annual Meeting of the Association for Computational Linguistics (Volume 1: Long Papers)},
  month     = jul,
  year      = {2026},
  address   = {San Diego, California, United States},
  publisher = {Association for Computational Linguistics},
  pages     = {23557--23580},
  doi       = {10.18653/v1/2026.acl-long.1081},
  url       = {https://aclanthology.org/2026.acl-long.1081/}
}

@article{ning2024userllm,
  title   = {User-{LLM}: Efficient {LLM} Contextualization with User Embeddings},
  author  = {Ning, Lin and Liu, Luyang and Wu, Jiaxing and Wu, Neo and Berlowitz, Devora and Prakash, Sushant and Green, Bradley and O'Banion, Shawn and Xie, Jun},
  journal = {arXiv preprint arXiv:2402.13598},
  year    = {2024},
  doi     = {10.48550/arXiv.2402.13598},
  url     = {https://arxiv.org/abs/2402.13598}
}

@inproceedings{liu-etal-2025-llms,
  title     = {{LLM}s + Persona-Plug = Personalized {LLM}s},
  author    = {Liu, Jiongnan and Zhu, Yutao and Wang, Shuting and Wei, Xiaochi and Min, Erxue and Lu, Yu and Wang, Shuaiqiang and Yin, Dawei and Dou, Zhicheng},
  booktitle = {Proceedings of the 63rd Annual Meeting of the Association for Computational Linguistics (Volume 1: Long Papers)},
  month     = jul,
  year      = {2025},
  address   = {Vienna, Austria},
  publisher = {Association for Computational Linguistics},
  pages     = {9373--9385},
  doi       = {10.18653/v1/2025.acl-long.461},
  url       = {https://aclanthology.org/2025.acl-long.461/}
}

@inproceedings{sun2020testtime,
  title     = {Test-Time Training with Self-Supervision for Generalization under Distribution Shifts},
  author    = {Sun, Yu and Wang, Xiaolong and Liu, Zhuang and Miller, John and Efros, Alexei and Hardt, Moritz},
  booktitle = {Proceedings of the 37th International Conference on Machine Learning},
  pages     = {9229--9248},
  year      = {2020},
  volume    = {119},
  series    = {Proceedings of Machine Learning Research},
  publisher = {PMLR},
  url       = {https://proceedings.mlr.press/v119/sun20b.html}
}

@inproceedings{sun2025learning,
  title     = {Learning to ({L}earn at Test Time): {RNN}s with Expressive Hidden States},
  author    = {Sun, Yu and Li, Xinhao and Dalal, Karan and Xu, Jiarui and Vikram, Arjun and Zhang, Genghan and Dubois, Yann and Chen, Xinlei and Wang, Xiaolong and Koyejo, Sanmi and Hashimoto, Tatsunori and Guestrin, Carlos},
  booktitle = {Proceedings of the 42nd International Conference on Machine Learning},
  pages     = {57503--57522},
  year      = {2025},
  volume    = {267},
  series    = {Proceedings of Machine Learning Research},
  publisher = {PMLR},
  url       = {https://proceedings.mlr.press/v267/sun25h.html}
}

@inproceedings{behrouz2025titans,
  title     = {Titans: Learning to Memorize at Test Time},
  author    = {Behrouz, Ali and Zhong, Peilin and Mirrokni, Vahab},
  booktitle = {Advances in Neural Information Processing Systems},
  volume    = {38},
  year      = {2025},
  doi       = {10.52202/085713-3786},
  url       = {https://proceedings.neurips.cc/paper_files/paper/2025/hash/a4ca07aa108036f80cbb5b82285fd4b1-Abstract-Conference.html}
}

@misc{tandon2025endtoend,
  title         = {End-to-End Test-Time Training for Long Context},
  author        = {Tandon, Arnuv and Dalal, Karan and Li, Xinhao and Koceja, Daniel and R{\o}d, Marcel and Buchanan, Sam and Wang, Xiaolong and Leskovec, Jure and Koyejo, Sanmi and Hashimoto, Tatsunori and Guestrin, Carlos and McCaleb, Jed and Choi, Yejin and Sun, Yu},
  year          = {2025},
  eprint        = {2512.23675},
  archivePrefix = {arXiv},
  primaryClass  = {cs.LG},
  url           = {https://arxiv.org/abs/2512.23675}
}

@inproceedings{chen2026perk,
  title     = {{PERK}: Long-Context Reasoning as Parameter-Efficient Test-Time Learning},
  author    = {Chen, Zeming and Romanou, Angelika and Weiss, Gail and Bosselut, Antoine},
  booktitle = {The Fourteenth International Conference on Learning Representations},
  year      = {2026},
  url       = {https://openreview.net/forum?id=qxDTe8fIyA}
}

@misc{kuratov2026gradmem,
  title         = {{GradMem}: Learning to Write Context into Memory with Test-Time Gradient Descent},
  author        = {Kuratov, Yuri and Kairov, Matvey and Bulatov, Aydar and Rodkin, Ivan and Burtsev, Mikhail},
  year          = {2026},
  eprint        = {2603.13875},
  archivePrefix = {arXiv},
  primaryClass  = {cs.CL},
  url           = {https://arxiv.org/abs/2603.13875}
}

@misc{lei2026deltamem,
  title         = {{$\delta$-mem}: Efficient Online Memory for Large Language Models},
  author        = {Lei, Jingdi and Zhang, Di and Li, Junxian and Wang, Weida and Fan, Kaixuan and Liu, Xiang and Liu, Qihan and Ma, Xiaoteng and Chen, Baian and Poria, Soujanya},
  year          = {2026},
  eprint        = {2605.12357},
  archivePrefix = {arXiv},
  primaryClass  = {cs.AI},
  doi           = {10.48550/arXiv.2605.12357},
  url           = {https://arxiv.org/abs/2605.12357}
}

@inproceedings{xiao-etal-2026-alpsbench,
  author    = {Jianfei Xiao and
               Xiang Yu and
               Chengbing Wang and
               Wuqiang Zheng and
               Xinyu Lin and
               Kaining Liu and
               Hongxun Ding and
               Yang Zhang and
               Wenjie Wang and
               Fuli Feng and
               Xiangnan He},
  title     = {{AlpsBench}: An {LLM} Personalization Benchmark for
               Real-Dialogue Memorization and Preference Alignment},
  booktitle = {Proceedings of the 49th International ACM SIGIR Conference
               on Research and Development in Information Retrieval},
  pages     = {3489--3497},
  year      = {2026},
  doi       = {10.1145/3805712.3808634},
  url       = {https://doi.org/10.1145/3805712.3808634}
}

@inproceedings{zhang-etal-2025-prime,
    title = "{PRIME}: Large Language Model Personalization with Cognitive Dual-Memory and Personalized Thought Process",
    author = "Zhang, Xinliang Frederick  and
      Beauchamp, Nicholas  and
      Wang, Lu",
    booktitle = "Proceedings of the 2025 Conference on Empirical Methods in Natural Language Processing",
    month = nov,
    year = "2025",
    address = "Suzhou, China",
    publisher = "Association for Computational Linguistics",
    url = "https://aclanthology.org/2025.emnlp-main.1711/",
    doi = "10.18653/v1/2025.emnlp-main.1711",
    pages = "33707--33736"
}

@inproceedings{zhao2024wildchat,
  title     = {{WildChat}: {1M} {ChatGPT} Interaction Logs in the Wild},
  author    = {Zhao, Wenting and Ren, Xiang and Hessel, Jack and Cardie, Claire and Choi, Yejin and Deng, Yuntian},
  booktitle = {The Twelfth International Conference on Learning Representations},
  year      = {2024},
  url       = {https://openreview.net/forum?id=Bl8u7ZRlbM}
}
\bibliographystyle{iclr2027_conference}

\appendix
\section{Appendix}
\subsection{History Serialization and Chunked TTT Updates}
\label{app:memory-writing-details}

\paragraph{Serializing the history.}
We write each user-authored record as a separate \texttt{user} turn in ChatML and keep the records in their original order. The memory is neither updated per token nor written from the full history in one pass. As in TTT-E2E \citep{tandon2025endtoend}, we split the serialized history into chunks of up to \(B=2{,}048\) tokens and apply a single TTT update to each chunk, which keeps the number of updates we have to differentiate through during meta-learning reasonable.

Records are packed into chunks greedily. If a record does not fit in the current chunk, we start a new one. We split a record only if it exceeds \(B\) tokens on its own. ChatML already marks turn boundaries, so we add no extra separators.

\paragraph{Per-chunk memory updates.}
The frozen LM goes through the chunks in order, each time using the memory obtained after the previous chunk. On each chunk, we compute the next-token prediction loss only over tokens of the original user records (ChatML role labels and turn boundaries are part of the input but not of the loss) and use it to update the memory parameters, and the next chunk is then processed with this updated memory. Serialization, masking, and the update procedure are identical in meta-learning and at inference.

\paragraph{ALPSBench.}
ALPSBench differs from the other sources in that its histories are complete user--assistant transcripts, not collections of user records. They are also short, and only about \(32\%\) of the tokens come from user turns, so writing the memory from user turns alone would leave too few tokens to work with. Moreover, the assistant responses in these transcripts often reflect preferences, experiences, and context that the user expressed in the preceding turns. For ALPSBench, we thus take the loss over both user and assistant content.

Each transcript is divided into consecutive user--assistant exchanges, and these exchanges are packed into \(2{,}048\)-token chunks. We keep the original ChatML role of every utterance, i.e., assistant responses are not relabeled as user-authored text. For all other training sources, the loss is still computed on user-authored content only.

\subsection{Training Details}
\label{app:training-details}

\paragraph{Training data construction.}
Each meta-learning example consists of a user history processed through TTT and a semantic-memory target used to supervise the shared memory initialization. Our training data combines user--assistant session dialogues from ALPSBench Tasks 1, 2, and 4 \citep{xiao-etal-2026-alpsbench} with posts and comments written by CMV users. Table~\ref{tab:meta-learning-data} summarizes these two data sources.

\begin{table}[t]
\caption{Composition of the meta-learning data.}
\label{tab:meta-learning-data}
\centering
\footnotesize
\setlength{\tabcolsep}{4pt}
\renewcommand{\arraystretch}{1.08}
\begin{tabularx}{\columnwidth}{@{}lXXr@{}}
\toprule
Source
& User history
& Meta-learning target
& \# examples \\
\midrule
ALPSBench
& User--assistant session dialogues
& Human-verified user-memory annotations
& 3,033 \\

CMV
& User-authored posts and comments
& Generated semantic user profiles
& 1,321 \\
\midrule
Total & & & 4,354 \\
\bottomrule
\end{tabularx}
\end{table}

For ALPSBench, we pool the development and validation partitions and construct a fixed 80/20 training--validation split at the session level. For Task 1, we pair each session dialogue with its complete set of verified memory items. We cast Task 2 in the same dialogue-to-profile format, creating two separate training examples: the original dialogue paired with its original memory, and the extended dialogue, which includes the new interaction, paired with the updated memory. Task 4 provides both a profile-extraction target and a request-conditioned target that asks the model to recall the memory item relevant to a new request.

For CMV, we use the training partition of ReCAP \citep{park-etal-2026-learning}. We retain users with at least eight history entries and use at most their \(100\) most recent posts and comments. For each user--post pair, GPT-5-mini generates a semantic profile from the user history and current post using the profile-generation prompt released with ReCAP. The profile describes user characteristics relevant to persuasion in the current discussion and serves as the meta-learning target.

\paragraph{Meta-learning prompt templates.}
For ALPSBench Tasks 1 and 2, as well as the profile-extraction target of Task 4, the model is prompted to recover the semantic information stored in the user memory:

\begin{quote}
\small
What do you remember about this user from your conversations with them?
List each fact you remember about the user, one per line.
\end{quote}

The target contains the corresponding verified memory items, rendered as one
bullet point per line. For the request-conditioned target in Task 4, we use:

\begin{quote}
\small
Here is a new request from this user:

\texttt{\{query\}}

What do you remember about this user that is relevant to this request?
State the relevant fact you remember.
\end{quote}

For CMV, the profile prompt additionally provides the current post:

\begin{quote}
\small
Here is a new post from this user:

\texttt{\{post\}}

What do you remember about this user that would help persuade them on this
post? List each fact you remember, one per line.
\end{quote}

The user history is not included in these prompts. It is processed separately through TTT, so the information required to generate each target must be obtained from the resulting user memory.

\paragraph{Profile-supervised meta-learning.}
We use Qwen3-0.6B as the frozen language-model backbone. For each training example, the memory parameters are initialized from the shared initialization and updated by applying TTT to the user history. The model then receives the corresponding meta-learning prompt and generates its target using the resulting memory.

We compute the teacher-forced next-token prediction loss over the target tokens and differentiate it through the complete sequence of TTT updates. The resulting gradient updates the shared memory initialization, while the language-model backbone remains frozen. When a training example contains multiple targets, as in Task 4, each target is processed separately, and their losses are combined according to the number of supervised target tokens.

\paragraph{Training configuration.}
The hyperparameters used to train the main model are reported in
Table~\ref{tab:training-hyperparameters}.

\begin{table}[t]
\caption{Hyperparameters used to train the main model.}
\label{tab:training-hyperparameters}
\centering
\small
\setlength{\tabcolsep}{4pt}
\renewcommand{\arraystretch}{1.08}
\begin{tabularx}{\columnwidth}{@{}p{0.43\columnwidth}X@{}}
\toprule
Hyperparameter & Value \\
\midrule
Backbone & Qwen3-0.6B \\
Memory rank & 18 \\
History chunk size & 2,048 tokens \\
TTT optimizer & SGD \\
TTT learning rate & 0.1 \\
TTT gradient clipping & 1.0 \\
Meta-learning optimizer & AdamW \\
Meta-learning rate & \(3\times10^{-4}\) \\
AdamW coefficients & \(\beta_1=0.9,\ \beta_2=0.95\) \\
Weight decay & 0.1 \\
Meta-gradient clipping & 1.0 \\
Learning-rate schedule
& 200-step warmup; cosine decay over 10,000 steps \\
Examples per meta-learning update & 2 \\
Source sampling & Proportional to \(\sqrt{N_s}\) \\
Training precision & bfloat16 \\
Training duration & 3 epochs (6,531 updates) \\
\bottomrule
\end{tabularx}
\end{table}

Each meta-learning update uses two examples drawn from distinct task-level sources. To prevent the largest source from dominating training, sources are sampled in proportion to the square root of their number of examples. We train the model for three epochs, corresponding to \(6{,}531\) meta-learning updates.

\subsection{Additional PersonaMem Results}
\label{app:additional-personamem}

We evaluate all seven categories of PersonaMem \citep{jiang-etal-2025-know}, which contains 589 questions with four answer choices each. Types 6 and 7 are the focus of our main evaluation. These questions require the model to draw on earlier interactions with a user to make a recommendation or respond to a new situation, making them particularly relevant to the semantic user memory we study. The remaining categories cover several other demands on memory: recalling specific facts and events, remembering the order in which preferences changed, and recognizing whether an item appeared in the history.

\begin{table}[t]
\centering
\caption{
\textbf{PersonaMem accuracy across all released question categories.}
\textit{Before} and \textit{after} denote performance without and with target-user information, respectively. For \textsc{ProTTT}, the two after-TTT rows differ only in the memory readout scale. Bold indicates the highest point estimate in each column; ties are all bolded. The number of questions is shown in parentheses.
\vspace{0.1cm}
}
\label{tab:personamem-all-types}
\setlength{\tabcolsep}{2.6pt}
\renewcommand{\arraystretch}{1.06}
\resizebox{\linewidth}{!}{%
\begin{tabular}{@{}lrrrrrrrr@{}}
\toprule
\textbf{Method}
& \shortstack{\textbf{Track}\\\textbf{Evolution}\\(139)}
& \shortstack{\textbf{Shared}\\\textbf{Facts}\\(129)}
& \shortstack{\textbf{Update}\\\textbf{Reasons}\\(99)}
& \shortstack{\textbf{New}\\\textbf{Ideas}\\(93)}
& \shortstack{\textbf{Mentioned}\\\textbf{Facts}\\(17)}
& \shortstack{\textbf{T6}\\\textbf{Recommend}\\(55)}
& \shortstack{\textbf{T7}\\\textbf{Generalize}\\(57)}
& \shortstack{\textbf{All}\\(589)}
\\
\midrule
Base
& .273 & .287 & .444 & .183 & \textbf{.353}
& .273 & .088 & .275
\\
\midrule
\multicolumn{9}{l}{\textit{In-Context Baselines}}
\\
\addlinespace[1pt]
Full History
& .230 & .287 & .313 & .172 & .294
& .273 & .140 & .244
\\
RAG (Recent-5)
& .295 & .318 & .525 & .151 & .235
& .418 & .140 & .311
\\
RAG (BM25)
& .259 & \textbf{.326} & .455 & .140 & .118
& \textbf{.491} & .246 & .304
\\
PAG
& .201 & .279 & .394 & .151 & .235
& .309 & .123 & .246
\\
PAG + Full History
& .237 & .318 & .323 & .172 & \textbf{.353}
& .273 & .123 & .255
\\
\midrule
\multicolumn{9}{l}{\textit{Parametric Baselines}}
\\
\addlinespace[1pt]
User-Specific LoRA
& .252 & .295 & .253 & .183 & .294
& .236 & .088 & .234
\\
\quad w/o user
& .281 & .287 & .434 & .194 & \textbf{.353}
& .309 & .088 & .280
\\
\addlinespace[2pt]
P2P
& .259 & .310 & .253 & .183 & \textbf{.353}
& .255 & .088 & .243
\\
\quad w/o user
& .259 & .310 & .253 & .183 & \textbf{.353}
& .255 & .088 & .243
\\
\addlinespace[2pt]
$\delta$-Mem
& .353 & .256 & \textbf{.616} & .151 & \textbf{.353}
& .309 & .140 & \textbf{.319}
\\
\quad w/o user
& .281 & .287 & .434 & .194 & \textbf{.353}
& .309 & .088 & .280
\\
\midrule
\multicolumn{9}{l}{\textit{Training Control}}
\\
\addlinespace[1pt]
TTT (naive)
& .201 & .287 & .364 & .172 & .294
& .273 & .140 & .246
\\
\quad w/o user
& .273 & .287 & .444 & .183 & \textbf{.353}
& .273 & .088 & .275
\\
\midrule
\multicolumn{9}{l}{\textit{Ours}}
\\
\addlinespace[1pt]
\rowcolor{protttmain}
\textsc{ProTTT} ($\alpha=.75$)
& \textbf{.367} & .279 & .404 & .129 & \textbf{.353}
& .364 & \textbf{.351} & .314
\\
\rowcolor{protttmain}
\textsc{ProTTT} ($\alpha=1$)
& .309 & .264 & .283 & .194 & .235
& .327 & .298 & .275
\\
\rowcolor{protttlight}
\quad w/o user
& .223 & .248 & .253 & \textbf{.247} & .059
& .218 & .246 & .234
\\
\bottomrule
\end{tabular}%
}
\end{table}

In these other categories, we find less consistent gains when answering correctly requires precise recall of the history. This fits the emphasis in \textsc{ProTTT} on retaining reusable knowledge about a user, with less attention to preserving a detailed record of past interactions. Some fine-grained details are retained, but the method is most useful when it can apply its broader knowledge of the user to a personalization task.

\subsection{Baseline Details}
\label{app:baseline-details}

\paragraph{Common evaluation protocol.}
All methods use Qwen3-0.6B as the prediction backbone and are evaluated using the official prompts and scoring protocol of each benchmark. We limit each user's history to the \(100\) most recent entries during both training and evaluation. With this limit, Full History can include a comparable amount of history without exceeding the model's context window. Context-based methods read the history as part of the prompt, while parametric methods first turn it into user-specific parameters or memory and then process the task input.

\paragraph{Full History, Recent-5, and BM25.}
Full History includes the user history directly in the prompt, while Recent-5 includes only the five most recent entries. BM25 retrieves five entries from the history using the current task input as the query. The Base condition receives the task input alone.

\paragraph{User-Specific LoRA.}
Following the input-only personalization setting examined by PRIME \citep{zhang-etal-2025-prime}, we train a separate LoRA adapter for each user through next-token prediction on their history. The backbone remains frozen, and the adapted model receives only the task query during evaluation. We apply rank-\(16\) LoRA with a scaling factor of \(32\) to all attention and feed-forward layers. The one-epoch configuration is selected using validation data and applied across all benchmarks. The corresponding \textit{w/o user} condition uses the unmodified base model.

\paragraph{Profile-Augmented Generation.}
Following the profile-augmented personalization paradigm of \citet{richardson-etal-2023-integrating}, PAG summarizes the retained user history into a textual profile and prepends the profile to each task input. We use GPT-5-mini to generate a query-independent profile once for each user and reuse it across subsequent queries. The same profile-generation prompt is used for all benchmarks:

\begin{quote}
\small
\textbf{System:} You write concise user profiles. Given messages written by
one user, you distill who they are.

\medskip
\textbf{User:} Below are messages written by a single user, oldest first.

\medskip
\texttt{[USER HISTORY]}

\medskip
Write a concise profile of this user: their interests, recurring topics, opinions and values, reasoning style, and writing style. Use short bullet points. Include only what the messages support. Output only the bullet points.
\end{quote}

PAG provides only the generated profile, whereas PAG + Full History provides both the profile and the same history used by Full History.

\paragraph{\texorpdfstring{$\delta$}{delta}-Mem.}
We train \(\delta\)-Mem using the authors' official implementation and training recipe \citep{lei2026deltamem}. For a direct comparison with our method, we replace its original backbone with Qwen3-0.6B while retaining the original memory architecture, objective, and training procedure.

At evaluation time, we write the user history into memory using the method's original procedure. The model then processes the task input with this memory, without the history in the prompt. For the \textit{w/o user} condition, we leave the memory empty.

\begin{table}[t]
\caption{
\textbf{Configurations of the parametric baselines.}
}
\label{tab:baseline-configurations}
\centering
\small
\setlength{\tabcolsep}{4pt}
\renewcommand{\arraystretch}{1.08}
\begin{tabularx}{\columnwidth}{@{}lXX@{}}
\toprule
\textbf{Method}
& \textbf{User-specific parameters}
& \textbf{Training settings}
\\
\midrule
User-Specific LoRA
& Rank \(16\), scale \(32\), attention and feed-forward projections
& AdamW, lr \(10^{-3}\), one epoch
\\
P2P
& Hypernetwork-generated rank-\(8\) LoRA on query and value projections
& AdamW, lr \(2\times10^{-5}\), profile-supervised training
\\
\(\delta\)-Mem
& Rank-\(8\) token-wise memory on query and output projections
& AdamW, lr \(2\times10^{-4}\), one QASPER epoch
\\
\bottomrule
\end{tabularx}
\end{table}

\subsection{Efficiency Measurement Details}
\label{app:efficiency-details}

\paragraph{Setup.}
We measure efficiency on the test sets of the main experiments: the 74 personalized queries of the 12 CMV test users, one query each from 100 randomly sampled LaMP-3 test users, and the 61 PersonaMem Type~6 and~7 queries, which come from 16 user-memory groups. All methods use the Qwen3-0.6B backbone and run on a single NVIDIA H200 NVL in bfloat16 with batch size one, using at most the 100 most recent history entries.

We report two costs separately: the one-time cost of turning a user's history into a user representation, and the cost paid on every query. PAG profiles are generated through the external GPT-5-mini API, so their construction cost is not reported. A query is one forward pass over the task prompt. We time each query five times after one untimed warm-up, take the median per query, and report the median over queries. Query latency includes everything a method repeats per query. For BM25, this covers retrieval, including building the index. For P2P, it covers retrieval, embedding the profile and retrieved history, generating the adapter with the hypernetwork, and attaching it. KV-cache size is computed from the prompt length, and peak memory is the maximum allocated GPU memory. GFLOPs are counted with the PyTorch FLOP counter, with the attention term halved for causal masking; for P2P we add the analytic cost of Qwen3-Embedding-4B.

With a 0.6B model, a short prompt needs little GPU compute, so latency is dominated by CPU-side kernel launches and follows the host CPU clock. For LaMP-3 and PersonaMem, we pin the process to one CPU core and keep only measurements taken while that core runs at its turbo frequency. This keeps 41--100 of the 100 LaMP-3 queries, depending on the method, and nearly all PersonaMem queries. History processing is timed without pinning. The No Meta-Learning control shares the architecture and memory-writing procedure of \textsc{ProTTT} and is not measured separately.

\paragraph{CMV.}
Table~\ref{tab:query-efficiency-details} gives the query-time costs, and Table~\ref{tab:state-construction-details} gives the cost of building each user representation. User-specific LoRA, $\delta$-Mem, and \textsc{ProTTT} build their parameters once and reuse them for every query. P2P generates a textual profile with Qwen2.5-7B-Instruct, then for every query generates a 1.15M-parameter adapter from the profile and retrieved history using Qwen3-Embedding-4B and a 126.47M-parameter hypernetwork. Profile generation takes 5.43\,s per user, and each adapter takes 0.063\,s; the adapter cost is counted in query latency.

\begin{table}[t]
\centering
\caption{
\textbf{Query-time cost on CMV.}
Prompt length and KV-cache size are reported as mean/max, and latency as median/mean.
P2P latency includes query-specific adapter generation.
For P2P, the second peak-memory value corresponds to adapter generation and is measured with \texttt{nvidia-smi} because it runs in a separate process.
}
\label{tab:query-efficiency-details}
\scriptsize
\setlength{\tabcolsep}{2.2pt}
\renewcommand{\arraystretch}{1.08}
\resizebox{\linewidth}{!}{%
\begin{tabular}{@{}lrrrrr@{}}
\toprule
\textbf{Method}
& \shortstack{\textbf{Prompt Tokens}\\mean / max}
& \shortstack{\textbf{KV Cache (MiB)}\\mean / max}
& \shortstack{\textbf{Peak Memory}\\(MiB)}
& \shortstack{\textbf{Latency (ms)}\\median / mean}
& \shortstack{\textbf{GFLOPs}\\per query}
\\
\midrule
Base
& 748 / 1,583
& 82 / 173
& 1,383
& 15.3 / 17.3
& 732
\\
Full History
& 18,919 / 29,544
& 2,069 / 3,231
& 5,166
& 273.5 / 218.7
& 63.6K
\\
RAG (Recent-5)
& 2,223 / 7,023
& 243 / 768
& 2,119
& 20.0 / 23.7
& 2.7K
\\
RAG (BM25)
& 4,255 / 14,189
& 465 / 1,552
& 3,088
& 39.2 / 49.2
& 6.9K
\\
PAG
& 1,066 / 1,931
& 117 / 211
& 1,430
& 15.9 / 18.0
& 1.1K
\\
PAG + Full History
& 19,215 / 29,927
& 2,102 / 3,273
& 5,217
& 278.1 / 222.0
& 65.3K
\\
User-Specific LoRA
& 748 / 1,583
& 82 / 173
& 1,526
& 29.6 / 29.5
& 747
\\
P2P
& 748 / 1,583
& 82 / 173
& 1,387 / 12,581
& 84.7 / 84.8
& 11.9K
\\
$\delta$-Mem
& 748 / 1,583
& 82 / 173
& 1,397
& 31.0 / 30.9
& 734
\\
\textsc{ProTTT}
& 748 / 1,583
& 82 / 173
& 1,432
& 26.2 / 26.1
& 733
\\
\bottomrule
\end{tabular}%
}
\end{table}

\begin{table}[t]
\centering
\caption{
\textbf{User representations on CMV and the cost of constructing them.}
Time is reported as mean/max per user. PAG profiles are generated through the external GPT-5-mini API, so their construction time and GPU memory are not included. P2P adapter generation is performed for each query and is included in the query latency reported in Table~\ref{tab:query-efficiency-details}; its profile-generation peak memory is measured using \texttt{nvidia-smi}.
}
\label{tab:state-construction-details}
\scriptsize
\setlength{\tabcolsep}{3.0pt}
\renewcommand{\arraystretch}{1.08}
\resizebox{\linewidth}{!}{%
\begin{tabular}{@{}llcrr@{}}
\toprule
\textbf{Method}
& \textbf{User Representation}
& \textbf{Construction}
& \shortstack{\textbf{Time (s)}\\mean / max}
& \shortstack{\textbf{Peak Memory}\\(MiB)}
\\
\midrule
Full History
& Raw history (17.8K tokens)
& --
& --
& --
\\
RAG (Recent-5)
& Five most recent entries
& --
& --
& --
\\
RAG (BM25)
& Raw history
& --
& --
& --
\\
PAG
& 268-token profile
& Generate profile (GPT-5-mini)
& n/a
& n/a
\\
PAG + Full History
& Raw history + 268-token profile
& Generate profile (GPT-5-mini)
& n/a
& n/a
\\
User-Specific LoRA
& 10.09M parameters
& One NTP epoch
& 11.74 / 13.05
& 17,450
\\
P2P
& Profile + 1.15M parameters/query
& Generate profile (Qwen2.5-7B)
& 5.43 / 7.00
& 21,727
\\
$\delta$-Mem
& 1,792 scalars
& Token-wise memory writing
& .29 / .47
& 2,816
\\
\textsc{ProTTT}
& 1.06M parameters
& TTT memory writing
& 2.82 / 4.50
& 8,263
\\
\bottomrule
\end{tabular}%
}
\end{table}

\paragraph{LaMP-3 and PersonaMem.}
Table~\ref{tab:additional-efficiency} reports the same comparison on the other two benchmarks.

\begin{table}[t]
\centering
\caption{
\textbf{Efficiency on LaMP-3 and PersonaMem.}
History processing is the mean one-time cost per user for LaMP-3 or memory group for PersonaMem. PAG profiles are generated through the external GPT-5-mini API, so their construction time is not reported. For P2P, history processing includes profile generation only; query-specific adapter generation is included in query latency. Prompt length is averaged over queries, and latency is the median.
}
\label{tab:additional-efficiency}
\small
\setlength{\tabcolsep}{5pt}
\renewcommand{\arraystretch}{1.06}
\begin{tabular*}{\linewidth}{@{\extracolsep{\fill}}lrrr@{}}
\toprule
\textbf{Method}
& \shortstack{\textbf{History Proc.}\\(s/user) $\downarrow$}
& \shortstack{\textbf{Prompt Tokens}\\/query $\downarrow$}
& \shortstack{\textbf{Latency}\\(ms/query) $\downarrow$}
\\
\midrule
\multicolumn{4}{l}{\textit{LaMP-3}}
\\
\addlinespace[1pt]
Base
& -- & 170 & 14.2
\\
Full History
& -- & 12.5K & 90.3
\\
RAG (Recent-5)
& -- & 750 & 14.9
\\
RAG (BM25)
& -- & 1.17K & 18.6
\\
PAG
& n/a & 404 & 14.9
\\
PAG + Full History
& n/a & 12.7K & 98.4
\\
User-Specific LoRA
& 11.85 & 170 & 28.8
\\
P2P
& 5.33 & 170 & 87.8
\\
$\delta$-Mem
& .20 & 170 & 30.8
\\
\textsc{ProTTT}
& 1.18 & 170 & 15.1
\\
\midrule
\multicolumn{4}{l}{\textit{PersonaMem Types 6 and 7}}
\\
\addlinespace[1pt]
Base
& -- & 394 & 14.3
\\
Full History
& -- & 10.5K & 100.7
\\
RAG (Recent-5)
& -- & 1.03K & 15.4
\\
RAG (BM25)
& -- & 1.05K & 17.9
\\
PAG
& n/a & 634 & 15.2
\\
PAG + Full History
& n/a & 10.8K & 104.7
\\
User-Specific LoRA
& 9.61 & 394 & 28.5
\\
P2P
& 6.53 & 394 & 76.7
\\
$\delta$-Mem
& .21 & 394 & 30.0
\\
\textsc{ProTTT}
& .89 & 394 & 14.2
\\
\bottomrule
\end{tabular*}
\end{table}

\paragraph{Summary.}
On all three benchmarks, \textsc{ProTTT} answers with the same task-only prompt as the base model (748, 170, and 394 tokens on average), while Full History puts the whole history into every prompt (18.9K, 12.5K, and 10.5K tokens). Its query latency stays close to the base-model level (26.2, 15.1, and 14.2\,ms, against 15.3, 14.2, and 14.3\,ms), compared with 273.5, 90.3, and 100.7\,ms for Full History. User-specific LoRA and $\delta$-Mem take about twice as long per query as \textsc{ProTTT} on LaMP-3 and PersonaMem, while the difference is smaller on CMV. P2P takes three to six times as long because it generates a new adapter for every query. Writing the memory takes 2.82, 1.18, and 0.89\,s and grows with the amount of history (22.2K, 13.9K, and 10.4K tokens on average). On every benchmark, this is less than what user-specific LoRA and P2P spend building their representations; only $\delta$-Mem writes faster. PAG is excluded from this comparison because its profiles are generated through an external API.

\subsection{Meta-Learning Supervision Objectives}
\label{app:meta-learning-objective-details}

We compare four meta-learning supervision objectives: profile prediction, persuasion judgment, response selection, and response generation. The models are trained on the same CMV users and differ only in the task they perform after memory construction.

Training examples come from the CMV split of \citet{park-etal-2026-learning}. Each example contains a user's post, their earlier posts and comments, and the replies to the post. On CMV, a post author awards a delta ($\Delta$) to a reply that changes their view. For each example, we write the user's 100 most recent history entries into memory before presenting the task instruction. We exclude examples with fewer than eight history entries. Because the history is never included in the task prompt, the model must draw on memory for information about the user. Table~\ref{tab:outer-objective-construction} gives the exact prompts for all four tasks.

\paragraph{Profile prediction.} The model describes what it remembers about the user that could help persuade them about the post. The reference profile is written by GPT-5-mini using the post, the same 100 history entries, and the profile-generation prompt from \citet{park-etal-2026-learning}. GPT-5-mini reads the history directly; the model being trained must recover the relevant information from memory.

\paragraph{Persuasion judgment.} The model sees one reply and judges whether it would change the user's view. The target answer is yes for a persuasive reply and no for a non-persuasive one. We use up to five replies of each type per post, sampling randomly when more are available.

\paragraph{Response selection.} The model identifies the reply that changed the post author's view among four replies to the same post: one persuasive and three non-persuasive, presented in shuffled order. The model answers with the number of the persuasive reply. Each post yields up to three questions, each built around a different persuasive reply. We exclude posts with fewer than three non-persuasive replies.

\paragraph{Response generation.} The model writes a reply that would change the user's view. Its training target is one of the post's persuasive replies, selected at random when several are available.

\begin{table}[t]
\caption{
Prompts and training targets of the four meta-learning tasks.
[POST] is the user's new post, [RESPONSE] is one reply to it, and [OPTIONS] lists the four candidate replies, numbered (1)--(4).
}
\label{tab:outer-objective-construction}
\centering
\footnotesize
\setlength{\tabcolsep}{3.5pt}
\renewcommand{\arraystretch}{1.10}
\begin{tabularx}{\linewidth}{
@{}
>{\raggedright\arraybackslash}p{0.17\linewidth}
>{\raggedright\arraybackslash}X
>{\raggedright\arraybackslash}X
@{}
}
\toprule
\textbf{Objective}
& \textbf{Training prompt}
& \textbf{Target} \\
\midrule

Profile prediction
&
\emph{Here is a new post from this user: [POST].
What do you remember about this user that would help persuade them on this post?
List each fact you remember, one per line.}
&
Bullet-point profile written by GPT-5-mini from the post and the user's history.
\\

Persuasion judgment
&
\emph{Here is a new post from this user: [POST].
And here is a response to it: [RESPONSE].
Would this response change this user's view?
Answer with just ``yes'' or ``no''.}
&
\texttt{yes} for a persuasive reply, \texttt{no} otherwise.
\\

Response selection
&
\emph{Here is a new post from this user: [POST].
Here are four responses to it: [OPTIONS].
Which response would change this user's view?
Answer with just the number.}
&
Number of the persuasive reply.
\\

Response generation
&
\emph{Here is a new post from this user: [POST].
Write the response that would most likely change this user's view.}
&
One persuasive reply to the post.
\\

\bottomrule
\end{tabularx}
\end{table}
\subsection{Continual-Memory Stream Construction and Evaluation}
\label{app:continual-construction}

\paragraph{Held-out ALPSBench users.}
We use ALPSBench Task~2 users from the held-out portion of the session-level split used for meta-learning. None of these users appear during meta-learning. We keep records with both the original and updated dialogues, a non-empty gold profile, and a change between the original and updated profiles. For each user, we concatenate the two dialogues into a single history segment and use the updated profile as the reference description. We keep segments that contain between \(2{,}000\) and \(9{,}000\) tokens after serialization.

\paragraph{Constructing compatible user streams.}
To simulate a longer user history in which new user characteristics appear over time, we combine history segments from multiple users. GPT-5-mini screens candidate profiles and groups users whose characteristics are compatible without being substantially redundant. We do not modify the original dialogues or profile statements.

For the information-acquisition experiment, we construct \(29\) disjoint streams of four users. Their history segments are written in \(\mathrm{A}\rightarrow\mathrm{B}\rightarrow\mathrm{C}\rightarrow\mathrm{D}\) order.

For the retention experiment, we construct eight-user streams, written in \(\mathrm{A}\rightarrow\cdots\rightarrow\mathrm{H}\) order. We use the eight streams whose complete histories fit within the context limit needed for the Full History comparison. Each stream begins independently from the shared memory initialization.

\paragraph{Sequential memory writing.}
We write the history segments in each stream to memory one at a time. Each TTT update processes only the newly added segment, starting from the memory produced by the previous update; earlier segments are not replayed. History serialization, chunking, and TTT updates follow Appendix~\ref{app:memory-writing-details}. We use the default memory scale, \(\alpha=1\), throughout memory writing and evaluation.

\paragraph{Scoring user information.}
After each write, we measure the mean token log-probability assigned to each user's gold profile statements. We evaluate each statement independently through teacher forcing under the following prompt:

\begin{quote}
\small
What do you remember about this user from your conversations with them?
List each fact you remember about the user, one per line.
\end{quote}

We first average the token log-probabilities within each statement, then average the resulting scores across that user's statements. Higher scores mean that the model assigns greater probability to the corresponding user information. Gold profiles are used only for evaluation and are never provided during memory writing.

\paragraph{Acquiring new user information.}
Figure~\ref{fig:continual-user-memory}(a) shows whether the memory acquires the characteristics associated with each newly added history segment. For each profile, we subtract its score under the shared initialization before any history has been written. We then average the change at each stage across the \(29\) four-user streams. A positive change indicates that the corresponding user information has become more likely under the progressively constructed memory.

\paragraph{Retaining earlier user information.}
Figure~\ref{fig:continual-user-memory}(b) tracks information associated with A as seven additional histories are written. For each stream, we use A's score immediately after writing A as the reference and measure how this score changes after writing B through H. We average the changes across the eight streams. The value at A is therefore zero, while subsequent values indicate how much of A's initially acquired information remains after later writes.

For Full History, we accumulate the same history segments directly in the prompt in A--H order and use the prompt containing only A as the reference. Full History and \textsc{ProTTT} are evaluated on the same eight streams. For \textsc{ProTTT}, the previously written histories are not included in the profile-evaluation prompt.

These synthetic streams contain compatible and complementary user information. They test whether the memory can acquire newly introduced characteristics while retaining previously acquired knowledge, but do not test contradictory changes in a single user's preferences.
\subsection{Generation and User-Information Recall Details}
\label{app:generation-analysis-details}

We test generation quality with user memory attached and ask whether the model can describe the user from information stored in that memory.

\paragraph{WildChat prompt set.}
Our generation set consists of \(40\) prompts sampled from WildChat after filtering and deduplication. We restrict the set to English prompts containing \(8\) to \(150\) words, removing those with URLs, code, or substantial non-text content. The selected prompts contain no information about the CMV users used in this analysis.

\paragraph{Generation protocol.}
For each of the \(9\) CMV test users from the general-capability analysis, we construct memory by having \textsc{ProTTT} write the user's \(100\) most recent history entries. We write this memory once per user and reuse it for all \(40\) prompts. Each personalized model is evaluated with \(\alpha=0.75\) and \(\alpha=1.0\), and we include the frozen base model for comparison.

We sample five responses to each prompt at temperature \(0.3\), using a no-repeat-4-gram constraint and a maximum of \(2{,}048\) generated tokens. User history is absent from the generation prompt. With five responses to each of the \(40\) prompts, we obtain \(200\) responses per user and \(1{,}800\) in total at each memory scale. We also generate \(200\) responses from the base model on the same prompts. Since the base model has no user-specific state, we run it only once.

\paragraph{Generation metrics.}
We use four metrics to measure non-termination, repetition, and diversity. For metrics based on words or n-grams, we tokenize on whitespace.

\begin{itemize}
    \item \textbf{Non-termination} is the fraction of responses that reach the \(2{,}048\)-token generation limit without terminating.
    \item \textbf{Sequential repetition-2} is the fraction of generated bigrams that duplicate another bigram in the same response. Lower values indicate less repetition within a response.
    \item \textbf{Distinct-2} is the ratio of unique bigrams to all generated bigrams. Higher values indicate greater lexical diversity.
    \item \textbf{Self-BLEU} measures similarity among responses from the same model. Lower values indicate greater diversity across responses.
\end{itemize}

For the memory conditions, we compute each metric separately for each user and average across the \(9\) personalized models. We compute Self-BLEU using BLEU-4 with smoothing: each response serves as a hypothesis, and the remaining responses from the same model serve as references. For this metric, we use a fixed subsample of \(100\) responses per model.

\paragraph{Generating user information from memory.}
To test whether the model can express stored user information, we ask it to describe each of the same \(9\) CMV users with the following prompt:

\begin{quote}
\small
What do you know about this user?
Describe their interests, opinions, and personality, being as specific as you can.
\end{quote}

The prompt includes neither user history nor a textual profile. For each user and model setting, we generate five responses at temperature \(0.3\), with a no-repeat-4-gram constraint and a limit of \(512\) generated tokens. This yields \(45\) user descriptions per setting.

\paragraph{Claim-level evaluation.}
GPT-5-mini splits each description into distinct claims and checks each claim against the corresponding user's history. It assigns exactly one label to each claim:

\begin{itemize}
    \item \textbf{Supported}: the history provides clear evidence for the claim;
    \item \textbf{Generic}: the claim contains no user-specific information;
    \item \textbf{Unsupported}: the claim is user-specific but lacks support in the history; or
    \item \textbf{Contradicted}: the claim directly conflicts with the history.
\end{itemize}

Table~\ref{tab:generation-quality}(b) reports the mean number of claims per description and the mean number assigned to each category.

\subsection{\textsc{ProTTT} with Llama-3.2-1B-Instruct}
\label{app:additional-backbone}

We further evaluate \textsc{ProTTT} with Llama-3.2-1B-Instruct to examine whether the proposed memory architecture and profile-supervised meta-learning procedure remain effective with a different backbone. We use the same training sources, objective, and TTT procedure as in the main experiments. Rank-$18$ memory modules are attached to all $16$ Transformer layers.

\begin{table}[t]
\centering
\caption{
\textbf{Personalization performance and efficiency with Llama-3.2-1B-Instruct.}
\textit{w/o user} denotes the shared initialization before constructing memory from the target user's history. Efficiency is measured on CMV using a single NVIDIA H200 NVL. History processing is a one-time per-user cost, while prompt length and latency are measured per query. $^\dagger$PAG profiles are generated offline using GPT-5-mini through an external API; profile-construction latency is therefore excluded from the
H200 measurements. Bold indicates the best personalization result for each metric.
}
\label{tab:llama-backbone-results}

\setlength{\tabcolsep}{3.0pt}
\renewcommand{\arraystretch}{1.06}
\resizebox{\linewidth}{!}{%
\begin{tabular}{
@{}l
rrrrr
!{\vrule width 0.8pt}
rrr
@{}
}
\toprule
&
\multicolumn{5}{c}{\textbf{Personalization}}
&
\multicolumn{3}{c}{\textbf{Efficiency on CMV}}
\\
\cmidrule(lr){2-6}
\cmidrule(lr){7-9}

\textbf{Method}
& \shortstack{\textbf{CMV}\\AUC $\uparrow$}
& \shortstack{\textbf{LaMP-3}\\MAE $\downarrow$}
& \shortstack{\textbf{LaMP-3}\\RMSE $\downarrow$}
& \shortstack{\textbf{PMem}\\T6 $\uparrow$}
& \shortstack{\textbf{PMem}\\T7 $\uparrow$}
& \shortstack{\textbf{History Proc.}\\(s/user) $\downarrow$}
& \shortstack{\textbf{Prompt}\\(tokens/query) $\downarrow$}
& \shortstack{\textbf{Latency}\\(ms/query) $\downarrow$}
\\
\midrule

Base
& .598 & 1.889 & 2.347 & .255 & .088
& -- & 763 & 8.5
\\

\midrule
\multicolumn{9}{l}{\textit{In-Context Baselines}}
\\
\addlinespace[1pt]

Full History
& .515 & 1.745 & 2.146 & .255 & .088
& -- & 18.8K & 220.7
\\

RAG (Recent-5)
& .549 & 1.711 & 2.107 & .255 & .088
& -- & 2.2K & 14.6
\\

RAG (BM25)
& .555 & 1.709 & 2.066 & .255 & .088
& -- & 4.3K & 32.6
\\

PAG
& .583 & 1.761 & 2.104 & .255 & .088
& n/a$^\dagger$ & 1.1K & 10.2
\\

PAG + Full History
& .506 & 1.613 & 2.013 & .255 & .088
& n/a$^\dagger$ & 19.0K & 225.0
\\

\midrule
\multicolumn{9}{l}{\textit{Ours}}
\\
\addlinespace[1pt]

\rowcolor{protttmain}
\textsc{ProTTT}
& \textbf{.599}
& \textbf{.827}
& \textbf{1.150}
& .255
& \textbf{.421}
& 2.67
& 763
& 25.0
\\

\rowcolor{protttlight}
\quad w/o user
& .483
& 1.502
& 2.211
& \textbf{.291}
& .298
& --
& 763
& 27.2
\\

\bottomrule
\end{tabular}%
}
\end{table}

Table~\ref{tab:llama-backbone-results} shows that constructing memory from the user history substantially improves \textsc{ProTTT} over its shared initialization on CMV, LaMP-3, and PersonaMem Type~7. The resulting model matches or outperforms every context-based baseline across all five personalization metrics, with particularly large gains on LaMP-3 and PersonaMem Type~7. The effectiveness of profile-supervised user memory construction is therefore not limited to the Qwen backbone used in the main experiments.

The Llama results also retain the efficiency advantage of parametric user memory. After a one-time memory-construction cost of $2.67$ seconds per user, \textsc{ProTTT} processes the same $763$-token task prompt for every subsequent query. Compared with Full History, it uses approximately $24.6\times$ fewer prompt tokens and achieves $8.8\times$ lower query latency. Although the additional memory modules introduce some overhead relative to the plain base model, their recurring cost remains substantially lower than repeatedly processing the complete user history.

\paragraph{General capability, generation quality, and user-information recall.}
We next examine how the Llama-based memory affects general capability and generation, and whether the stored user knowledge can be expressed in generated text. We compare the full memory readout with a scaled readout, $\alpha=0.5$. The complete readout-scale sweep is reported in Appendix~\ref{app:readout-scale-sweep}.

\begin{table}[t]
\centering
\caption{
\textbf{General capability, generation quality, and user-information recall with Llama-3.2-1B-Instruct.}
(a) Reports 5-shot MMLU accuracy after processing $100$ history items and generation metrics on $40$ WildChat prompts.
(b) Reports the mean number of claims per generated user profile.
Sup., Gen., Unsup., and Contr. denote supported, generic, unsupported, and contradicted claims, respectively.
}
\label{tab:llama-generation-quality}

\small
\setlength{\tabcolsep}{3pt}
\renewcommand{\arraystretch}{1.10}

\resizebox{\linewidth}{!}{%
\begin{tabular}{@{}lrrrrr!{\vrule width 0.8pt}ccccc@{}}
\toprule
&
\multicolumn{5}{c}{\textbf{(a) General capability and generation quality}}
&
\multicolumn{5}{c}{\textbf{(b) Claims in generated user profiles}}
\\
\cmidrule(lr){2-6}
\cmidrule(l){7-11}

\textbf{Method}
& \shortstack{\textbf{MMLU}\\\textbf{Acc.} $\uparrow$}
& \shortstack{\textbf{Non-}\\\textbf{term.} $\downarrow$}
& \shortstack{\textbf{Seq.-}\\\textbf{rep.-2} $\downarrow$}
& \shortstack{\textbf{Distinct-2}\\$\uparrow$}
& \shortstack{\textbf{Self-}\\\textbf{BLEU} $\downarrow$}
& \shortstack{\textbf{Claims}\\\textbf{/gen.}}
& \textbf{Sup.}
& \textbf{Gen.}
& \textbf{Unsup.}
& \textbf{Contr.}
\\
\midrule

Base
& \textbf{.455}
& .000
& .071
& .609
& .304
& 8.88
& 3.93
& 2.13
& 2.62
& .20
\\

\midrule

\rowcolor{protttmain}
\textsc{ProTTT}, $\alpha=.5$
& .405
& .014
& .068
& .642
& .263
& 14.85
& \textbf{6.85}
& 2.38
& 5.27
& .35
\\

\rowcolor{protttlight}
\textsc{ProTTT}, $\alpha=1$
& .259
& .059
& .143
& .487
& .289
& 11.98
& 5.15
& .67
& 5.77
& .40
\\

\bottomrule
\end{tabular}%
}
\end{table}

Table~\ref{tab:llama-generation-quality}(a) shows that using the memory at full strength noticeably affects both MMLU and WildChat generation. Reducing the readout scale to $\alpha=0.5$ mitigates this effect: MMLU accuracy increases from $.259$ to $.405$, non-termination decreases from $5.9\%$ to $1.4\%$, and repetition and diversity remain close to or better than those of the base model.

The scaled memory also retains stronger access to user knowledge. As shown in Table~\ref{tab:llama-generation-quality}(b), $\alpha=0.5$ produces $6.85$ history-supported claims per generated profile, compared with $5.15$ at $\alpha=1$ and $3.93$ for the base model. Both memory settings also generate more unsupported claims than the base model, indicating that greater access to user knowledge does not uniformly improve factual precision. Overall, readout scaling provides a substantially better balance between general model behavior and the ability to verbalize semantic user knowledge.

\paragraph{Continual acquisition and retention.}
We also repeat the continual-memory analysis with Llama-3.2-1B-Instruct to examine whether the memory can acquire newly revealed user knowledge while retaining knowledge constructed from earlier history.

\begin{figure}[t]
    \centering
    \includegraphics[
        width=\linewidth,
        keepaspectratio
    ]{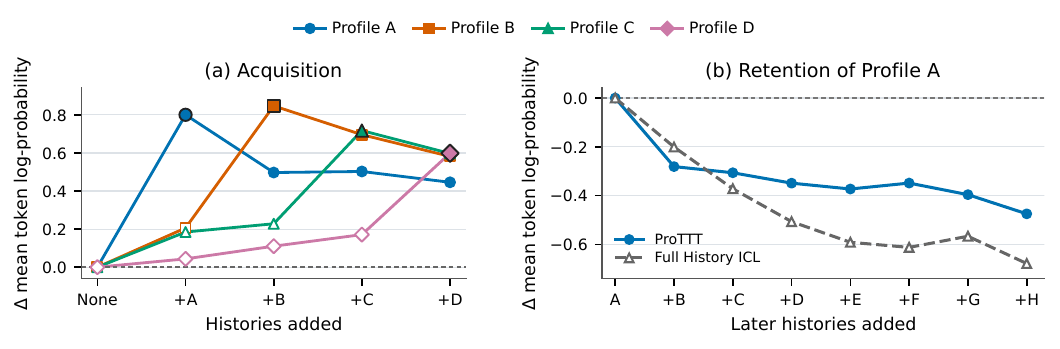}
    \caption{
    \textbf{Continual acquisition and retention with Llama-3.2-1B-Instruct.}
    (a) Change in the mean token log-probability of each gold user profile as the corresponding histories are sequentially incorporated.
    (b) Change in the mean token log-probability of Profile~A as seven later histories are incorporated, comparing \textsc{ProTTT} with full history ICL.
    Smaller degradation indicates better retention of previously acquired user knowledge.
    }
    \label{fig:llama-continual-user-memory}
\end{figure}

Figure~\ref{fig:llama-continual-user-memory}(a) shows that the mean token log-probability of each profile increases substantially when its corresponding history is incorporated. All four profiles remain above their initial values after the complete sequence, showing that the Llama-based memory can continually acquire newly revealed semantic user knowledge.

Figure~\ref{fig:llama-continual-user-memory}(b) examines whether previously acquired knowledge remains accessible as additional history accumulates. After seven later histories, the mean token log-probability of Profile~A decreases by approximately $.48$ under \textsc{ProTTT}, compared with approximately $.68$ under Full History. Although successive updates still cause some interference, \textsc{ProTTT} retains more of the earlier user knowledge than keeping the accumulated history directly in context. Together, these results reproduce the continual acquisition and retention behavior observed with the main Qwen backbone.

\subsection{Effect of the Memory Readout Scale}
\label{app:readout-scale-sweep}

The memory readout scale $\alpha$ determines the magnitude of the memory residual added to the frozen backbone. Memory construction is the same in every setting: we always process the user history with $\alpha=1$, producing identical memory parameters. We vary $\alpha$ only when applying that memory to subsequent predictions. This sweep thus isolates the effect of readout strength while keeping the memory-writing procedure and its computational cost fixed.

We evaluate $\alpha\in\{0.25,0.5,0.75,1.0\}$ with Qwen3-0.6B and Llama-3.2-1B-Instruct. We measure personalization on CMV, LaMP-3, and PersonaMem Types~6 and~7. We also evaluate 5-shot MMLU and WildChat generation after constructing memory from the same user histories used in the corresponding backbone experiments.

\begin{figure}[t]
    \centering
    \includegraphics[
        width=\linewidth,
        keepaspectratio
    ]{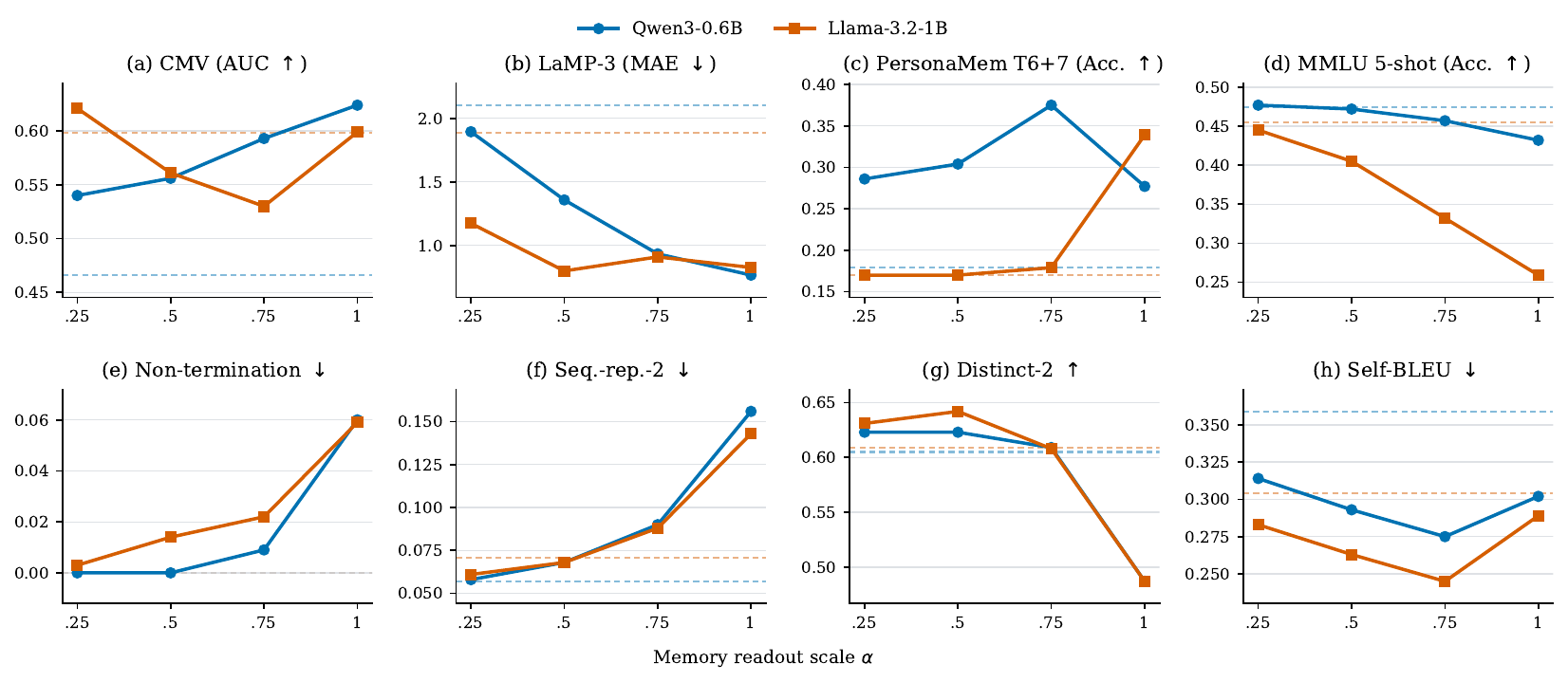}
    \caption{
    \textbf{Effect of the memory readout scale across backbone models.}
    Panels (a)--(c) show personalization performance on CMV, LaMP-3, and PersonaMem Types~6 and~7.
    Panel (d) shows 5-shot MMLU accuracy.
    Panels (e)--(h) show generation quality on WildChat.
    Solid lines represent \textsc{ProTTT} after user memory has been constructed; dashed lines represent the corresponding base model without user memory.
    Memory construction is identical across readout scales.
    }
    \label{fig:readout-scale-sweep}
\end{figure}

\paragraph{Qwen3-0.6B.}
Figures~\ref{fig:readout-scale-sweep}(a)--\ref{fig:readout-scale-sweep}(c) show that a stronger memory readout generally helps on CMV and LaMP-3. CMV AUC rises as $\alpha$ increases, while LaMP-3 MAE steadily falls. PersonaMem shows a different pattern: its combined Type~6 and~7 accuracy peaks at $\alpha=0.75$, rather than at full readout. The effect of readout strength therefore depends on the semantic user knowledge needed for the downstream task.

The MMLU and WildChat results show how readout strength affects general model behavior. At $\alpha=0.25$ and $\alpha=0.5$, MMLU stays close to the base model, and the WildChat metrics show little degradation in generation. As the scale increases, personalization improves, but MMLU accuracy gradually falls and repetition and non-termination increase. The largest shift occurs at $\alpha=1$: CMV and LaMP-3 reach their best results, while generation moves further from the base model.

\paragraph{Llama-3.2-1B-Instruct.}
The Llama results depend more strongly on the readout scale. CMV performs best at $\alpha=0.25$, LaMP-3 at $\alpha=0.5$, and PersonaMem Types~6 and~7 at $\alpha=1$. In contrast to Qwen, increasing $\alpha$ does not consistently help across the personalization benchmarks.

The decline in general performance is also more pronounced for Llama. MMLU remains close to the base model at $\alpha=0.25$, then falls as the readout scale increases. On WildChat, repetition and diversity remain close to the base model at $\alpha=0.25$ and $\alpha=0.5$. At full readout, non-termination and repetition increase, while Distinct-2 decreases.

Taken together, these results show that $\alpha$ controls how much the constructed user memory shapes model behavior. Smaller values tend to preserve the frozen backbone's behavior, while larger values can yield stronger personalization when the task benefits from a more influential memory. The best balance depends on both the task and the backbone; no single scale performs best across all evaluations.
\end{document}